%% file: main.tex
\pdfoutput=1

\documentclass[11pt]{article}

\usepackage[preprint]{acl}

\usepackage{times}
\usepackage{latexsym}

\usepackage[T1]{fontenc}

\usepackage[utf8]{inputenc}

\usepackage{microtype}

\usepackage{inconsolata}

\usepackage{graphicx}
\usepackage{amsmath}
\usepackage{booktabs}
\usepackage{adjustbox}
\usepackage{caption}
\usepackage{tcolorbox}
\usepackage[nameinlink]{cleveref}
\usepackage{float}
\usepackage{gensymb}
\usepackage{stfloats} 

\usepackage{xcolor} 

\title{What Are the Odds? \\Language Models Are Capable of Probabilistic Reasoning}

\author{%
 Akshay Paruchuri\thanks{Work completed during an internship at Google.}, Jake Garrison, Shun Liao, John Hernandez,\\ \textbf{Jacob Sunshine, Tim Althoff, Xin Liu, Daniel McDuff\textsuperscript}\\
 Google \\
 akshay@cs.unc.edu, \{althoff, xliucs, dmcduff\}@google.com
}

\begin{document}
\maketitle

\input{p_main_sections/0-abstract}
\input{p_main_sections/1-introduction}
\input{p_main_sections/2-related_works}
\input{p_main_sections/3-probabilistic_reasoning_tasks}
\input{p_main_sections/4-probabilistic_reasoning_datasets}
\input{p_main_sections/5-experimental_setup}
\input{p_main_sections/6-experimental_results}
\input{p_main_sections/7-conclusion}

\input{p_main_sections/8-limitations}

\clearpage 

\bibliography{references}

\input{p_appendix_sections/0-table_of_contents}
\input{p_appendix_sections/1-idealized_distribution_prompts}
\input{p_appendix_sections/2-idealized_distribution_results_summaries}

\input{p_appendix_sections/3-real_world_distribution_prompts}
\input{p_appendix_sections/4-additional_experimental_results}
\input{p_appendix_sections/5-broader_impacts}

\end{document}

%% file: p_main_sections/0-abstract.tex
\vspace{-10cm} 
\begin{abstract}
Language models (LM) are capable of remarkably complex linguistic tasks; however, numerical reasoning is an area in which they frequently struggle. An important but rarely evaluated form of reasoning is understanding probability distributions. In this paper, we focus on evaluating the probabilistic reasoning capabilities of LMs using idealized and real-world statistical distributions. We perform a systematic evaluation of state-of-the-art LMs on three tasks: estimating percentiles, drawing samples, and calculating probabilities. We evaluate three ways to provide context to LMs 1) anchoring examples from within a distribution or family of distributions, 2) real-world context, 3) summary statistics on which to base a Normal approximation. Models can make inferences about distributions, and can be further aided by the incorporation of real-world context, example shots and simplified assumptions, even if these assumptions are incorrect or misspecified. To conduct this work, we developed a comprehensive benchmark distribution dataset with associated question-answer pairs that we have released publicly.
\end{abstract}

%% file: p_main_sections/1-introduction.tex
\section{Introduction}
\label{sec:intro}

\begin{figure}[ht!]
    \centering
   \includegraphics[width=\linewidth]{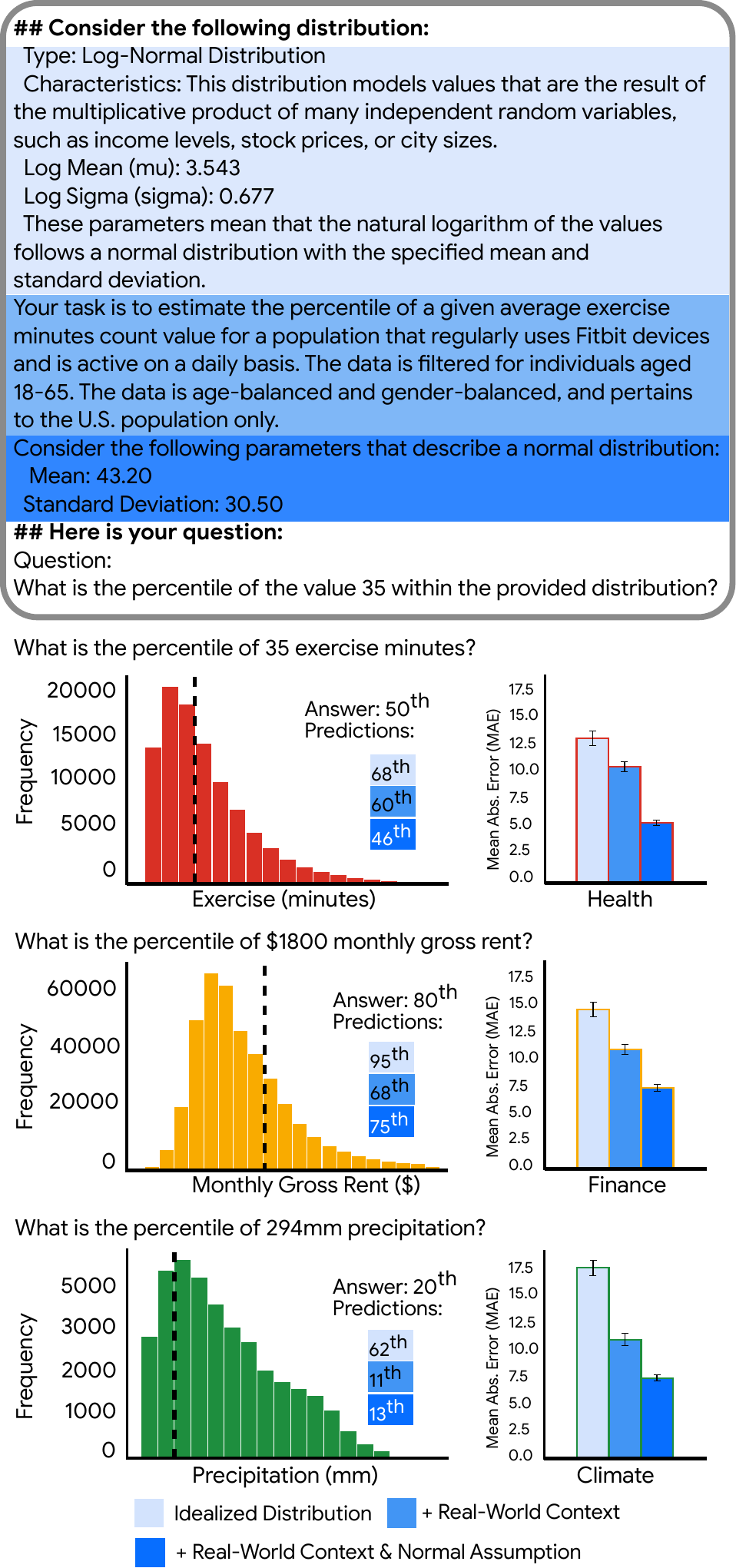}
    \caption{\textbf{LMs \& Probabilistic Reasoning.} 
    Models can make inferences about distributions, but can be aided by the incorporation of real-world context, example shots and simplified assumptions, even if these assumptions are incorrect or misspecified.
    }
    \vspace{-5mm}
    \label{fig:teaser}
\end{figure}

Language models (LMs)~\citep{workshop2022bloom, touvron2023llama, achiam2023gpt} are versatile interfaces to knowledge, capable of remarkably complex linguistic tasks. Summarization of complex documents~\cite{tang2023evaluating,zhang2024benchmarking}, reasoning over long passages of text~\cite{shaham2022scrolls, chen2023extending, team2023gemini} and zero-shot inference in specialist domains such as medicine~\cite{mcduff2023towards} are a few examples that demonstrate their abilities. While performance on primarily linguistic problems, can often be strong, effectiveness on operations that involve numerical reasoning is a domain that language models have struggled with~\citep{kojima2022large}. Difficulties handling numbers may be due to model pretraining formulations (e.g., using autoregressive next token prediction pretext tasks) or numerical token representations not necessarily being suited to mathematical reasoning~\cite{bachmann2024pitfalls}, or simply a limited representation of these types of tasks in the training corpora. Nevertheless, some work suggests prompting techniques can substantially improve LM performance on numerical reasoning tasks, indicating that relevant knowledge may already be encoded within these models~\citep{imani2023mathprompter}.

A form of numerical reasoning that is important for interpreting many different forms of data is contextualizing an individual measurement or measurements (a sample or samples) within a population (a distribution). Drawing insights from data frequently requires comparing and contrasting a sample from other samples. This is because absolute values in isolation can be hard to interpret, without the context of how probable they are or how close they are to the maximum or minimum values observed across the population. Probabilistic reasoning is something that the human brain appears to do~\cite{knill2004bayesian} and that is an important component in cognition~\cite{chater2006probabilistic}. Thinking probabilistically is efficient as one does not have to represent every detail of every sample that one observes, and instead can have the data summarized with a small number of parameters that describe the distribution~\cite{lindskog2021can}. Research has shown that some probabilistic reasoning processes lead to superior performance; for example, people are more accurate at answering questions about statistical properties when they estimate the full distribution first~\cite{goldstein2014lay}. Yet, there are limited examples evaluating or improving on the probabilistic reasoning by designing LMs that reason over sets~\cite{ozturkler2023thinksum}. 

Understanding the distributions is important in many contexts. In population level data it is important when gauging whether an individual behavior is normative (e.g., Is sleeping 8 hours normal for a college aged student?). In climatology, inferences about distributions of temperature or precipitation data on a given day of the year at a particular location are important when determining if observed events are typical or abnormal. Is a maximum temperature of 35\degree C likely to be observed in Seattle every year?

In this work we propose and define a set of probabilistic reasoning tasks and use them to evaluate the capabilities of LMs - \textit{estimating percentiles}, \textit{drawing samples}, and \textit{calculating probabilities}. 
Next we evaluate the impact of additional real-world context and parametric assumptions (Normal distribution) using the task of \textit{estimating percentiles} (task choice is motivated in~\Cref{sec:probabilistic_reasoning_tasks}).

To summarize our research questions:
\begin{enumerate}
    \item \Cref{sec:experimental_setup_idealized_distributions}: Provided with an idealized distribution, are language models able to accurately answer questions about them? Does this vary by the distribution family? Does providing prompt examples from different distributions in the same family or samples from the same distribution help? Do LMs simply repeat the nearest in-context example or is there evidence of more complex LM behavior?
    \item \Cref{sec:experimental_results_real_world_distributions}: Can an LM answer questions about distributions in the world (e.g., income in the US population)? Are LMs able to retrieve statistics and answer questions about these distributions in a zero-shot manner?
    \item \Cref{sec:experimental_results_real_world_distributions}: Using simple approximations such as assuming a Normal distribution, can we design prompts that lead to more accurate answers to probabilistic reasoning questions? 
\end{enumerate}

To answer these questions we develop a distribution dataset with associated question-answer pairs that we will release publicly. The dataset includes questions about 12 families of standard, idealized distributions (e.g., Normal or Power-law distributions) and distributions of real-world data from the domains of population health, climate, and finance. Code and additional results for our work can be found here: \url{https://github.com/yahskapar/LLMs-and-Probabilistic-Reasoning}.

%% file: p_main_sections/2-related_works.tex
\section{Related Work}

\noindent\textbf{Language Models and Numerical Reasoning.} Working with numbers is necessary for many everyday tasks. Yet, while large language models pretrained on vast numbers of documents exhibit impressive linguistic capabilities, they often struggle at tasks involving numerical reasoning~\cite{saxton2019analysing,kojima2022large} (for a survey see~\citet{lu2022survey}). Different approaches to numerical reasoning have been proposed, many focusing on logical reasoning of mathematical tasks~\cite{geva2020injecting,imani2023mathprompter,yang2022logicsolver,webb2023emergent}. 
In quantitative reasoning problems such as those in the domains of mathematics, science, and engineering, the process of fine-tuning models has been used to successfully remedy weaknesses~\cite{lewkowycz2022solving}. Automatic generation of data can be used as a way of obtaining training examples~\cite{geva2020injecting,liu2022generated}.  The fact that specific prompting, such as providing examples, can improve the performance of LMs on numerical tasks, suggests that their training data may already include relevant information to perform these tasks~\cite{imani2023mathprompter,yang2022logicsolver}. Benchmark datasets have helped the research community to develop these methods (e.g.,~\cite{he2023solving,zhang2024careful,liu2024llms}) further and automatic evaluation of numerical reasoning problems has been proposed to help in cases where accuracy cannot be computed mathematically~\cite{cobbe2021training}.

\noindent\textbf{Numerical Reasoning Prompt Design.} Prompts designed to handle automatically generated content (from an LM) were used to improve on numerical and scientific commonsense reasoning tasks~\cite{liu2022generated}. Algorithms and code are useful tools when working with numbers, chain-of-thought prompts have been designed to leverage these specifically to improve the performance of LMs on arithmetic problems~\cite{imani2023mathprompter, merrill2024transforming}.    Retrieval of correlated-examples~\cite{yang2022logicsolver}, generating intermediate reasoning steps~\cite{gao2023pal} and expressing reasoning as a program~\cite{chen2022program} are examples that can also help improve LM performance on mathematical and logic problems. Simple approaches such as zero-shot chain-of-thought~\cite{kojima2022large} exist and are capable of leveraging multi-step reasoning, but for probabilistic reasoning tasks lead to severely degraded performance due to generally poor numerical reasoning performance.

\noindent\textbf{Probabilistic Reasoning and Cognition.} Inspiration for AI systems is often drawn from our understanding of human cognition, cognitive science has revealed insights about how humans can think probabilistically~\cite{cosmides1996humans,oaksford2001probabilistic} and can build representations of relatively complex probability distributions~\cite{lindskog2021can}, yet our perceptions of means and variances are subject to biases~\cite{tversky1974judgment}. The thought processes people use when answering questions about distributions have an impact on their accuracy. Specifically, eliciting a full distribution before computing summary or sample statistics can make answers more accurate~\cite{goldstein2014lay}.

%% file: p_main_sections/3-probabilistic_reasoning_tasks.tex
\section{Defining Probabilistic Reasoning Tasks}
\label{sec:probabilistic_reasoning_tasks}

Our probabilistic reasoning benchmark contains three distinct tasks that explore a language model's (LM's) context-free (idealized) understanding of basic, \emph{idealized distributions}, we describe these tasks as follows:
\\
 \textbf{Task 1: Estimating Percentiles.} Given a distribution, the model is asked for the percentile a sample would appear in.  A question is composed of a value (a sample) that is calculated given the target percentile. The answer is expected to be a numerical response from 0 to 100. The target percentiles utilized were $n^{st/th} = \{1, 10, 20, 30, 40, 50, 60, 70, 80, 90, 99\}$. Language models responses to these questions are sampled 10 times with a random seed.
\begin{tcolorbox}[colback=black!5!white,colframe=black!75!black]
{\textit{Return the percentile of the value \{X\} within a normal distribution with mean \{Y\} and standard deviation \{Z\}.}}
\end{tcolorbox}

\noindent\textbf{Task 2: Drawing Samples.} Given a distribution the model is asked to draw samples at random from it. A random seed is used for each sample and we repeat this 1000 times per distribution. The language model is explicitly instructed to avoid generating any code or using additional tools to perform the sampling. The answer is expected to be a numerical response.

\begin{tcolorbox}[colback=black!5!white,colframe=black!75!black]
{\textit{Sample a number from the normal distribution with mean \{X\} and standard deviation \{Y\}.}}
\end{tcolorbox}

\noindent\textbf{Task 3: Calculating Probabilities.} Given a distribution the model is asked for the probability a sample from the distribution will fall between two given values.
The target probabilities and the corresponding target ranges are computed based on a lower and upper quantile to form examples with different probabilities in the set $\mathcal{P} = \{0.1, 0.2, 0.3, 0.4, 0.5, 0.6, 0.7, 0.8, 0.9, 1.0\}$. The answer from the LM is expected to be a numerical response from 0 to 1.

\begin{tcolorbox}[colback=black!5!white,colframe=black!75!black]{\textit{Calculate the probability that a value falls between \{W\} and \{X\} in a normal distribution with mean \{Y\} and standard dev. \{Z\}.}}
\end{tcolorbox}

In order to evaluate the impact of real-world context on LM probabilistic reasoning, we explore distributions in the real-world that have additional real-world context (e.g., prior knowledge and expectations about US household incomes). We then leverage Task 1 of \emph{estimating percentiles} to evaluate to what degree real-world context impacts performances. Task 1 is particularly well suited for this exploration, as we later demonstrate that this Task 1 elicits the highest variation in performance across distribution families and the number of in-context examples/shots (further detailed in~\Cref{sec:experimental_results}).
In~\Cref{sec:appendix_idealized_distribution_prompts} we provide full prompt templates for our three proposed benchmark tasks, as well as further details regarding template components and real-world distribution prompt templates.

\begin{figure*}[t!]
    \centering
    \includegraphics[width=1\textwidth]{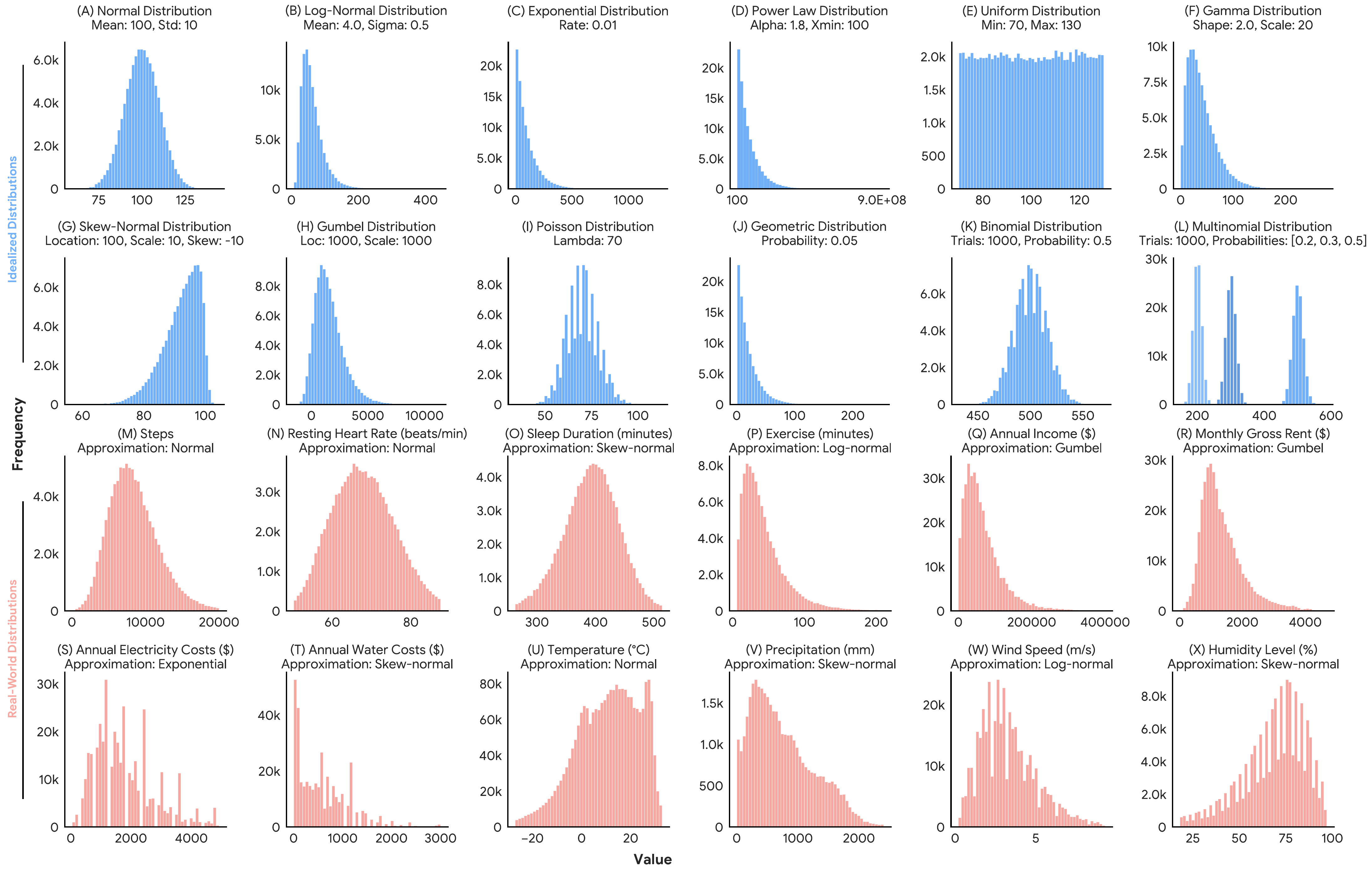}
    \vspace{-2em}
    \caption{\textbf{Distributions.} A visualization of the 12 idealized and 12 real-world distributions across the domains of health, finance, and climate involved in our evaluation.}
    \label{fig:all_distributions_full}
    \vspace{-0.4cm}
\end{figure*}

%% file: p_main_sections/4-probabilistic_reasoning_datasets.tex
\section{Curating Data Distributions}
\label{sec:probabilistic_reasoning_datasets}

We use two distinct datasets for the purpose of understanding the probabilistic reasoning capabilities of LMs - a dataset of \emph{idealized distributions} and a dataset of \emph{real-world distributions}. A visualization of both datasets is shown in~\Cref{fig:all_distributions_full}.

\noindent\textbf{Idealized Distributions.}
\label{sec:probabilistic_reasoning_datasets_idealized_distributions}
We identify 12 families of distributions: Normal, Log-Normal, Skew-Normal, Exponential, Power Law, Uniform, Gamma, Gumbel, Poisson, Geometric, Binomial and Multinomial. These encompass sets of distributions identified~\cite{frank2009common} and tested in prior studies~\cite{goldstein2014lay}. When an LM is asked about an idealized distribution, a \emph{distribution description} is provided with parameters that can range from simple parameters such as the mean and standard deviation (e.g., for a Normal distribution) to less easily interpretable parameters such as location and scale (e.g., for a Gumbel distribution). Note that the distribution description does not include any distrbution probability density function (PDF) or cumulative distribution function (CDF). All parameters are captured in a popular, public Python library for scientific computing - NumPy.

\noindent\textbf{Real-World Distributions.}
\label{sec:probabilistic_reasoning_datasets_real_world_distributions}
We choose real-world distributions from the domains of health, finance, and climate for which there is presumed to be relevant information in the model's training set.

\noindent\underline{Health:}
    We sample 100K Fitbit users from the U.S. population, aged 18-65, and with at least 10 days of data from the calendar year 2023. The dataset was gender and age balanced (see Appendix). We analyze four common wearable metrics \textbf{(A)} step count, \textbf{(B)} resting heart rate, \textbf{(C)} sleep duration, and \textbf{(D)} exercise minutes. We aggregate this data to obtain averages for each user and ultimately distributions of daily metrics across 100K users.
\begin{tcolorbox}[colback=black!5!white,colframe=black!75!black]
{\textit{In what percentile of the US population would someone with an average of 6000 steps per day be?}}
\end{tcolorbox}

\noindent\underline{Finance:} We use public census data from the Census Bureau’s American Community Survey (ACS) Public Use Microdata Sample (PUMS)~\cite{ruggles2020ipums} that contains measures of \textbf{(E)} annual income, \textbf{(F)} monthly gross rent, \textbf{(G)} annual electricity costs, and \textbf{(H)} annual water costs (\$) for individuals and households in the US. We select data from the calendar year 2018.

\begin{tcolorbox}[colback=black!5!white,colframe=black!75!black]
{\textit{In what percentile of US households would someone with a household income of \$70,000 be?}}
\end{tcolorbox}

\noindent\underline{Climate:} We use the public Global Historical Climatology Network daily (GHCNd) dataset~\cite{menne2012overview} maintained by the National Oceanic and Atmospheric Administration (NOAA) that contains average daily measures for variables of \textbf{(I)} temperature, \textbf{(J)} precipitation, \textbf{(K)} wind speed, and \textbf{(L)} humidity level for weather stations across the continental United States. We consider data from the calendar year 2018 and filter erroneous measures indicated by measurement quality flags built into the GHCNd dataset.

\begin{tcolorbox}[colback=black!5!white,colframe=black!75!black]
{\textit{In what percentile would an average temperature of 20 degrees Celsius be in the USA?}}
\end{tcolorbox}

\begin{figure*}[ht!]
    \centering
    \includegraphics[width=0.8\textwidth]{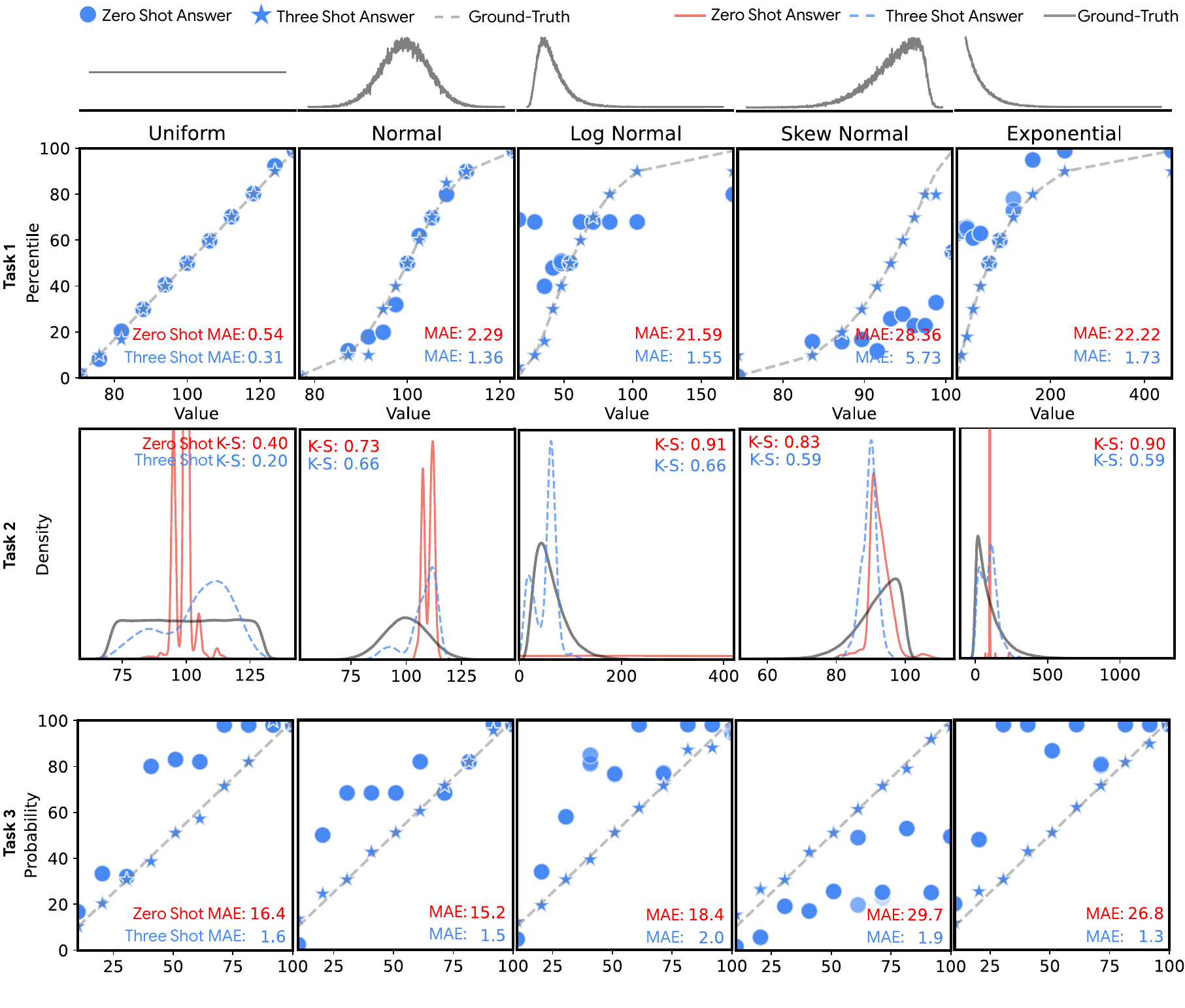}
    \caption{\textbf{Results on Idealized Distributions.} Model results (top) estimating percentiles, (middle) drawing samples, (bottom) estimating probabilities, for five common distributions (see~\Cref{sec:appendix_idealized_distributions_results_summaries} for results on all distributions).}
    \label{fig:synthetic_results}
    \vspace{-1.5em}
\end{figure*}

%% file: p_main_sections/5-experimental_setup.tex
\section{Experimental Setup}
\label{sec:experimental_setup}

\subsection{Idealized Distributions}
\label{sec:experimental_setup_idealized_distributions}

\noindent\textbf{Zero-shot Performance.} We evaluate the zero-shot performance of three LMs (\emph{Gemini 1.0 Ultra}, \emph{GPT4-Turbo}, and \emph{GPT3.5-Turbo}) across our three tasks and 12 idealized distributions. LM prompts are generated as formulated in~\Cref{sec:probabilistic_reasoning_tasks}.

\noindent\textbf{N-Shot Performance.} We propose two types of shots that might be reasonably employed for the tasks - \textit{within distribution family distribution shots} and \textit{within distribution shots}. \textit{Within distribution family distribution shots} are examples from a \underline{different distribution from the same family} as the current distribution in question. The distribution parameters of the variant are randomly sampled from a specified range of reasonable parameter values. For example, if we are asking for a percentile of a value in a normal distribution with a mean of 100 and a standard deviation of 10, and we are providing three shots, the randomized shots may be generated from three variant normal distributions with means of 108, 118, and 112 and corresponding standard deviations of 13, 16, and 10. \textit{Within distribution shots} entail shots from \underline{the same distribution} that is being asked about in a question.

To help contextualize the performance in the N-shot experiments, for the task of \emph{estimating percentiles} we compare both shot types to a baseline where the answer is picked based on the nearest corresponding target percentile value in the shot examples (i.e., the nearest neighbor). This baseline does not perform any interpolation between percentiles, which would be required for optimal performance. If the LM performance exceeds this baseline performance, it would suggest that the LMs does perform some form of interpolation, instead of simply reciting in-context examples. 

We explicitly avoid using shots that involve an answer that could be an answer to one of our proposed questions. Specifically, we use $n^{th}_{shots} = \{5, 15, 25, 35, 45, 55, 65, 75, 85, 95\}$ and $\mathcal{P}_{shots} = \{0.05, 0.15, 0.25, 0.35, 0.45, 0.55, 0.65, 0.75, 0.85, \newline0.95\}$. For example, if we are asking for a percentile of a value in a Normal distribution with a mean of 100 and a standard deviation of 10, and we are providing three shots, the mapped shots will be generated from the same normal distribution and correspond to 35.0, 55.0, and 75.0. 
We sample LM responses 10 times per question with a random seed. We provide further details, including examples per shot type and shot count, in our~\Cref{sec:appendix}.

\subsection{Real-World Distributions}
\label{sec:experimental_setup_real_world_distributions}

\noindent\textbf{Zero-shot Performance.} We evaluate the zero-shot performance of three LMs (\emph{Gemini 1.0 Ultra}, \emph{GPT4-Turbo}, and \emph{GPT3.5-Turbo}) across the proposed task of \textit{estimating percentiles} in order to evaluate an LM's understanding of probabilistic reasoning of the real-world distributions in the domains of health, finance, and climate mentioned in~\Cref{sec:probabilistic_reasoning_datasets_real_world_distributions}. We design prompts that contain information about the corresponding real-world data, such as where it was sourced from, the year in time for which the data is relevant, and any relevant filtering that was done, this acts as context that we refer to as ``Real-World Context'' in the remainder of the paper. The prompts are built using 11 unique target values that correspond to ground truth percentiles generated from the real-world data. We sample LM responses 10 times per question with a random seed.

\noindent\textbf{Performance by Context.} To investigate the effects of context, we compare two conditions: 1) questions about an idealized distribution with comparable shape and parameters of the real-world distribution but no other context, and 2) questions about the distribution with Real-World Context (see examples in Appendix~\ref{sec:appendix}). Both conditions involve a distribution description as mentioned in~\Cref{sec:probabilistic_reasoning_tasks}.
The prompts are again built using 11 unique target values that correspond to ground truth percentiles generated from the real-world data. We sample \emph{Gemini 1.0 Ultra} responses 10 times per question with a random seed. 

\noindent\textbf{Simplified Assumptions.} Certain real-world distributions found in~\Cref{sec:probabilistic_reasoning_datasets_real_world_distributions} are Non-Normal. For example, annual income follows a Power Law distribution. Despite not all distributions found in~\Cref{sec:probabilistic_reasoning_datasets_real_world_distributions} being perfectly Normal distributions, we devise a prompting strategy that involves treating any distribution in the prompt as a normal distribution with a specified mean and standard deviation. This approach can be further justified by the fact that, despite characteristics such as skewedness being present in real-world distributions, many distributions are similar to a Normal distribution. We quantitatively reinforce this observation using the Kolmogorov–Smirnov test~\cite{chakravarti1967handbook} to show that even if the Normal equivalent is not the best fit for a given real-world distribution, it can be remarkably close as evidenced by the K-S statistic. Additionally, we compare our proposed Normal approximation approach to simply providing a question involving a real-world distribution with three \emph{within distribution} shots.

%% file: p_main_sections/6-experimental_results.tex
\section{Experimental Results}
\label{sec:experimental_results}

We organize our results based on the research questions posed in~\Cref{sec:intro}. Alongside a concise answer in bold, we provide discussion based on our analysis of the experimental results.

\begin{table}[h!]
    \normalsize	
    \captionsetup{width=\columnwidth}
    \vspace{-8pt}
    \centering
    \normalsize	
    \setlength{\tabcolsep}{2pt}
    \adjustbox{max width=\columnwidth}{
    \begin{tabular}{rccc}
    \toprule[1.5pt]
        \textbf{Model} & \textbf{Percentiles (\%)} & \textbf{Sampling (K-S)} & \textbf{Probabilities (\%)} \\ 
        \midrule
        GPT3.5-Turbo & $25.7 \pm 3.11$ & $0.73 \pm 0.07$ & $32.7 \pm 2.38$ \\
        GPT4-Turbo & $\textbf{14.9}\pm \textbf{2.39}$ & $\textbf{0.59} \pm \textbf{0.08}$ & $21.0 \pm 2.11$ \\
        Gemini 1.0 Ultra & $16.5 \pm 2.67$ & $0.76 \pm 0.09$ & $\textbf{19.4} \pm \textbf{2.26}$ \\
        \bottomrule[1.5pt]
    \end{tabular}}
    \footnotesize
    \caption{\textbf{Aggregated zero-shot task performance across different LMs.} We evaluate zero-shot performance for tasks such as percentiles, sampling, and probabilities using Gemini 1.0 Ultra, GPT4-Turbo, and GPT3.5-Turbo. For the tasks of estimating percentiles and calculating probabilities, results are reported as Mean Absolute Error (MAE) $\pm$ Standard Error ($\sigma_{M}$). For the task of drawing samples, the Mean K-S statistic $\pm$ Standard Error ($\sigma_{M}$) is reported with all reported values having $p < 0.01$.}
    \label{tab:model_comparison_RQ1}
    \vspace{-0.2cm}
\end{table}

\subsection{Idealized Distributions}
\label{sec:experimental_results_idealized_distributions}

\noindent\textbf{Q1: Are LMs able to accurately answer questions about idealized distributions in a zero-shot setting?} \emph{Answer}: \textbf{It varies, performance on some idealized distributions is better than others.} \emph{Discussion}: Language model performance varied considerably across families of distribution. Zero-shot performance on the percentile task was best for the uniform (MAE = 0.54\%) and normal (MAE = 2.29\%) distributions (see ~\autoref{fig:synthetic_results} for examples and ~\autoref{fig:percentile_results_full} for more detailed results). Furthermore, the performance of LMs on answering questions about distributions varied by task. Estimating percentiles showed rather impressive zero-shot performance (MAE $\bar{\mu}$ = 16.52, min = 0.54, max = 28.36). Calculating probabilities was worse on average (MAE $\mu$ = 21.48, min = 9.03, max = 32.53) and sampling performance was generally poor in the zero-shot case. In all cases, providing shots (for different percentiles, samples or probabilities) \underline{within a distribution} improved the performance substantially (Percentile $\Delta_{0 \rightarrow 1 shot}$ = +59.14\%, Sampling $\Delta_{0 \rightarrow 1 shot}$ = +55.26\% K-S stat., Probability $\Delta_{0 \rightarrow 1 shot}$ = +70.13\%).

\begin{figure}[t]
    \centering
    \includegraphics[width=1\columnwidth]{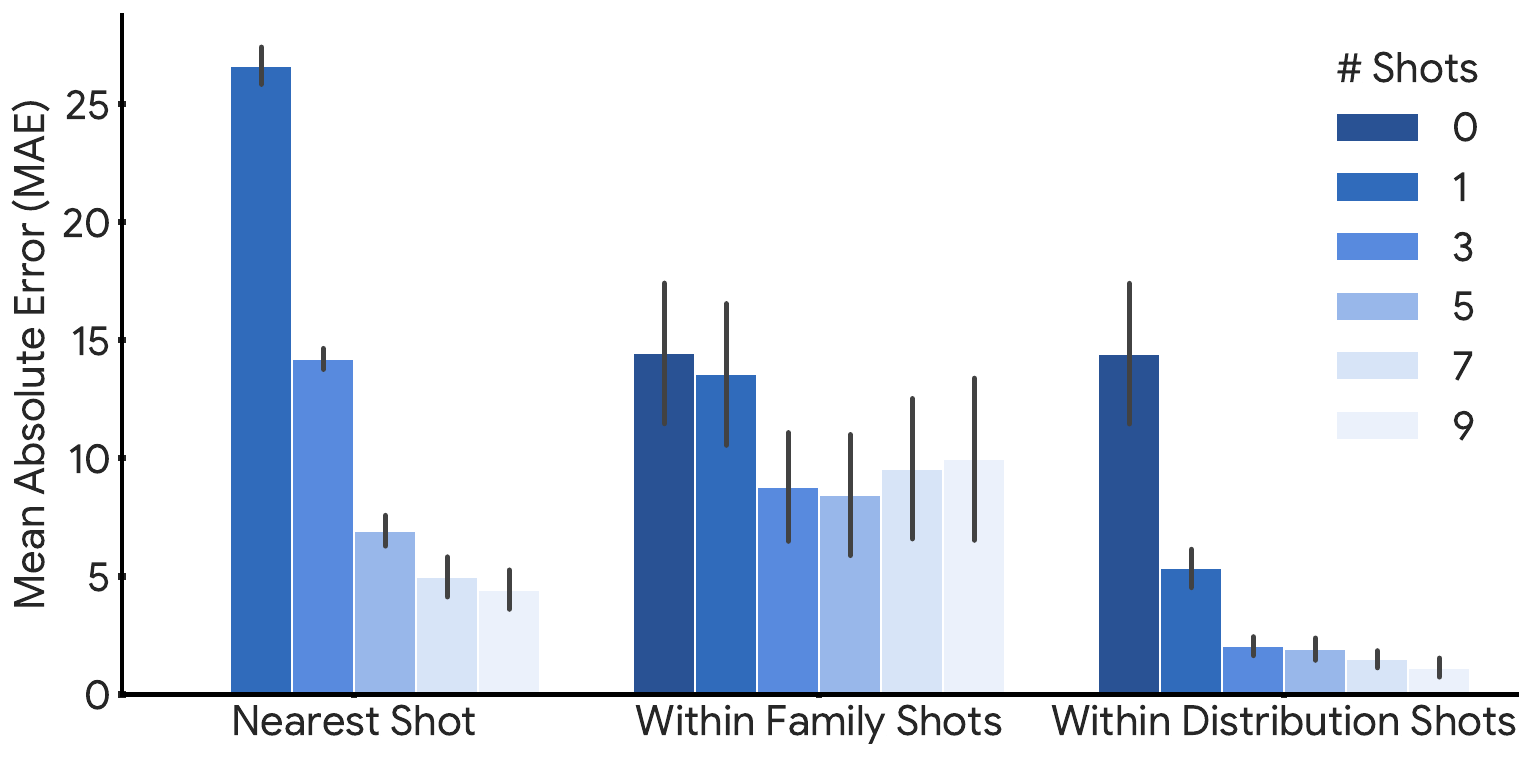}
    \caption{\textbf{Language models appear to interpolate between in-context examples.} 
    Comparison of within family and within distribution shot types to a baseline where the answer is based on the nearest corresponding shot to the target percentile value (nearest neighbor), importantly the baseline does not perform any interpolation between percentiles.}
    
    \label{fig:nearest_shot_experiment}
    \vspace{-1.5em}
\end{figure}

\noindent\textbf{Q2: Does providing prompt examples from different distributions in the same family or samples from the same distribution help?} \emph{Answer}: \textbf{Providing \emph{within distribution} shots helps more than \emph{within family} shots.} \emph{Discussion}: As illustrated in~\Cref{fig:nearest_shot_experiment}, providing prompt examples (shots) from different distributions from the same family has less impact on the performance than providing prompt examples from the same distribution in the question.

\noindent\textbf{Q3: Do LMs simply repeat the nearest in-context example?} \emph{Answer}: \textbf{No, they appear to perform some interpolation that is superior to a nearest-in-context baseline.} \emph{Discussion}: 
We observe that LM performance exceeds that of this baseline (~\Cref{fig:nearest_shot_experiment}) which suggests that LMs perform some kind of interpolation, instead of simply reciting in-context examples.

\begin{figure*}[t]
    \centering
    \includegraphics[width=1\textwidth]{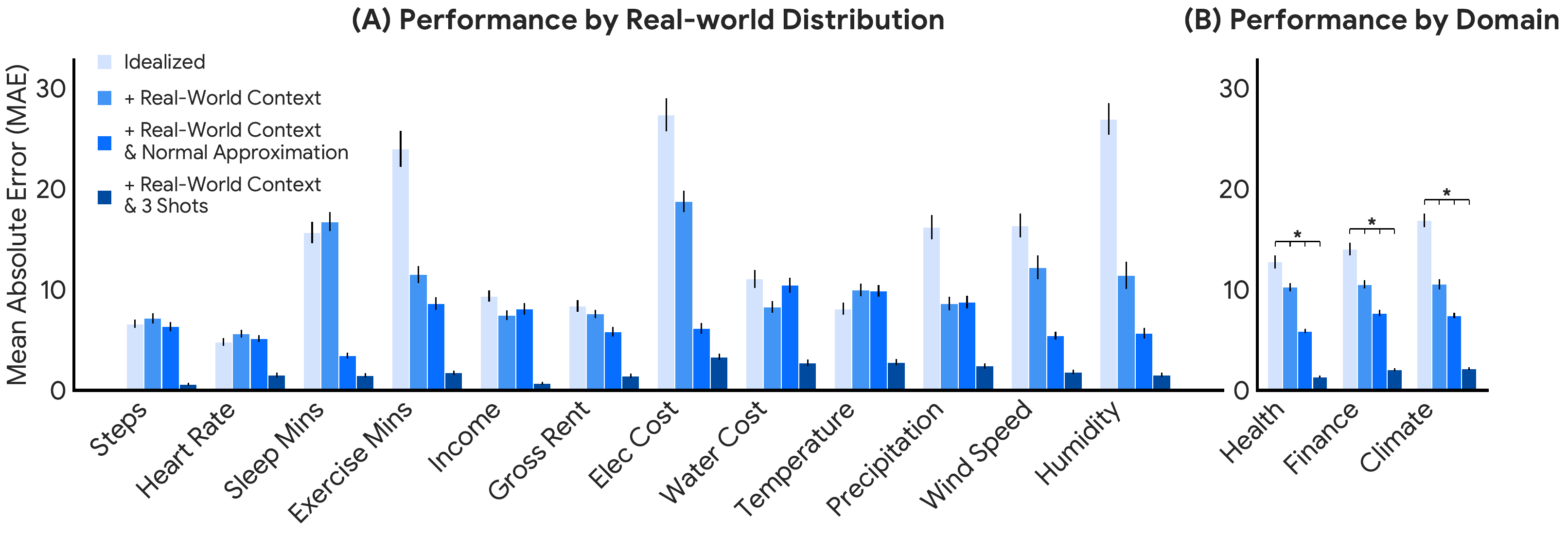}
    \vspace{-2em}
    \caption{\textbf{Inferences can be aided by context and simplified assumptions.} Mean absolute error in calculating percentiles for real-world distributions with different prompts, including idealized distributions without real-world context, added real-world context, and a Normal approximation approach that simplifies parameter content.
    (*) designates $p < 0.05$ for all possible pairs using the Wilcoxon signed-rank test.}
    \label{fig:realworld_results}
\end{figure*}

\begin{table*}[h!]
    \normalsize	
    \captionsetup{width=\textwidth}
    \vspace{-8pt}
    \centering
    \normalsize	
    \setlength{\tabcolsep}{2pt}
    \adjustbox{max width=1\textwidth}{
    \begin{tabular}{rccccccccc}
    \toprule[1.5pt]
        & \multicolumn{3}{c}{\textbf{Health}} & \multicolumn{3}{c}{\textbf{Finance}} & \multicolumn{3}{c}{\textbf{Climate}} \\
        \cmidrule(lr){2-4} \cmidrule(lr){5-7} \cmidrule(lr){8-10}
        \textbf{Model} & \textbf{Idealized} & \textbf{Real-world Con.} & \textbf{Norm. Approx.} & \textbf{Idealized} & \textbf{Real-world Con.} & \textbf{Norm. Approx.} & \textbf{Idealized} & \textbf{Real-world Con.} & \textbf{Norm. Approx.} \\ 
        \midrule
        GPT3.5-Turbo & $20.5 \pm 9.62$ & $20.3 \pm 8.51$ & $6.81 \pm 0.68$ & $17.7 \pm 4.54$ & $20.4 \pm 2.88$ & $7.55 \pm 0.77$ & $22.7 \pm 6.88$ & $25.7 \pm 6.32$ & $7.90 \pm 0.22$ \\
        GPT4-Turbo & $\textbf{11.0} \pm \textbf{4.94}$ & $\textbf{4.92} \pm \textbf{3.18}$ & $\textbf{3.15} \pm 0\textbf{0.76}$ & $\textbf{8.99} \pm \textbf{1.18}$ & $10.7 \pm 3.24$ & $\textbf{5.50} \pm \textbf{0.48}$ & $18.5 \pm 6.53$ & $15.2 \pm 5.13$ & $\textbf{4.94} \pm \textbf{0.58}$ \\
        Gemini 1.0 Pro & $25.30 \pm 8.41$ & $11.51 \pm 1.06$ & $10.42 \pm 1.32$ & $29.35 \pm 3.72$ & $11.77 \pm 0.92$ & $10.10 \pm 1.01$ & $26.20 \pm 5.44$ & $18.67 \pm 2.01$ & $16.53 \pm 1.94$ \\
        Gemini 1.0 Ultra & $12.8 \pm 4.43$ & $10.3 \pm 2.49$ & $5.89 \pm 1.09$ & $14.0 \pm 4.47$ & $\textbf{10.5} \pm \textbf{2.75}$ & $7.62 \pm 1.06$ & $\textbf{16.9} \pm \textbf{3.86}$ & $\textbf{10.5} \pm \textbf{0.79}$ & $7.43 \pm 1.11$ \\
        
        \bottomrule[1.5pt]
    \end{tabular}}
    \footnotesize
    \caption{\textbf{Zero-shot performance by domain and context category across different LMs.} All results are reported as Mean Absolute Error (MAE) $\pm$ Standard Error ($\sigma_{M}$) with (\%) units.}
    \label{tab:model_comparison_RQ2_RQ3}
    \vspace{-1.5em}
\end{table*}

\subsection{Real-World Distributions}
\label{sec:experimental_results_real_world_distributions}

\noindent\textbf{Q4: What is the zero-shot accuracy of an LM on distributions in the world (e.g., income in the US population)?} \emph{Answer}: \textbf{It varies, performance on some real-world distributions is better than others.} \emph{Discussion}: As shown in~\Cref{tab:model_comparison_RQ2_RQ3}, LMs are capable of varying degrees of zero-shot performance given different kinds of context. We consider the real-world context as the primary baseline in our investigation of real-world distributions, in contrast to idealized, context-free versions of the same real-world distributions and added context that simplifies assumptions (e.g., Normal approximation). On average with real-world context, \emph{Gemini 1.0 Ultra} has superior zero-shot performance on distributions in the climate domain. In contrast, \emph{GPT4-Turbo} has superior zero-shot performance in the health domain. Both \emph{Gemini 1.0 Ultra} and \emph{GPT4-Turbo} had comparable performance on the finance domain while being superior to \emph{GPT3.5-Turbo}. We attribute these differences in zero-shot performance to underlying differences in the large amounts of training data used to train LMs, especially in the case \emph{Gemini 1.0 Ultra} and \emph{GPT4-Turbo}.

\noindent\textbf{Q5: Does the provided real-world context help with probabilistic reasoning performance?} \emph{Answer}: \textbf{Yes.} \emph{Discussion}: ~\Cref{fig:realworld_results} details both distribution-wise and domain-wise results using \emph{Gemini 1.0 Ultra} with our four context categories - idealized, real-world context, real-world context with Normal approximation, and real-world context with 3 shots. On average, adding real world-context improves performance and adding real-world context with a Normal approximation improves performance still further. This trend is true on aggregate but not for all individual distributions (e.g., resting heart rate). This observation appears to be restricted to distributions that already have a reasonable baseline performance, we suspect that the saturated performance conflicts with the model's ability to leverage real-world context and simpler assumptions such as the Normal approximation. Additionally, certain distributions such as household income show a decrease in performance when Normal approximation is applied, likely because the household income distribution follows a Power Law distribution. It is unhelpful to apply a Normal approximation on distributions that differ greatly from a normal distribution. We empirically show this with the same set of 12 idealized distributions in~\Cref{sec:appendix_additional_normal_approximation_results}. Lastly, we note that real-world context with 3 shots has the best performance. This is unsurprising, and furthermore does not invalidate the impact of simplified assumptions such as Normal approximation, which can be more efficient due to not relying on 3 shots.

\noindent\textbf{Q6: Do parametric assumptions such as a Normal approximation as a prompt design strategy improve performance?} \emph{Answer}: \textbf{Yes.} \emph{Discussion}: On average across all domains, yes. The simple assumption of a Normal distribution performs well and when paired with real-world context, consistently improves performance on real-world distributions (see~\Cref{fig:realworld_results}). This seems reasonable given the aforementioned internal, potentially incorrect representations of real-world distributions that LMs can have, and subsequently how stats such as mean and standard deviation can help correct the LM's baseline knowledge. It is perhaps surprising that performance improves relatively consistently, despite the real-world distributions often differing from an idealized Normal distribution and therefore the LM is being conditioned on a misspecified, yet still helpful, model.

\noindent\textbf{Q7: How do simpler assumptions such as a Normal approximation compare to providing three few-shot examples?} \emph{Answer}: \textbf{Providing three few-shot examples is generally better.} \emph{Discussion}: Generally speaking, providing three few-shot examples is better and provides superior performance across our proposed domain-specific datasets of health, finance, and climate.

%% file: p_main_sections/7-conclusion.tex
\section{Conclusion}

LMs are able to answer questions about idealized and real-world distributions, with real-world results suggesting there is some internal representation that enables modeling or interpolation from distribution parameters. The probabilistic reasoning performance of LMs varies, with certain distributions (e.g., uniform, normal) having much better performance in contrast to other distributions (e.g., log-normal, skew-normal). Within distribution shots and context can improve probabilistic reasoning performance, as can a simplified Normal approximation.

%% file: p_main_sections/8-limitations.tex
\section{Limitations}

Numerical calculation and reasoning remains an area in which language models, even very large models, tend to perform poorly. Making approximations based on distributions is effective; however, it may also be a source of potential biases. Our experiments have not focused on a deep exploration of the ability of language models to represent and answer questions about extreme values, such as outliers in distributions. Our results do suggest that language model particularly struggle with accounting for extreme values in very skewed (long-tail) distributions. In computing percentiles the model would often overestimate the percentiles (and thus underestimate the presence of extreme values) - see the Power Law family results in Fig.~\ref{fig:percentile_results_full} of our appendices. Our work shows that, despite some promising zero-shot performance and ways to improve that performance, language models require more improvements with Non-Uniform and Non-Normal distributions before they are capable of being relied on for probabilistic reasoning of real-world distributions that follow other distributions (e.g., Power Law). We hope our insights as a part of this work, as well as our proposed tasks and datasets critical to probabilistic reasoning and to be publicly released, prove valuable to the community at large and their efforts to make language models more useful, safer, and ultimately more reliable.

%% file: p_appendix_sections/0-table_of_contents.tex
\clearpage 
\appendix

\section*{Overview of Appendices}
\label{sec:appendix}

The appendix is organized as follows:

\noindent\textbf{\Cref{sec:appendix_idealized_distribution_prompts}} contains additional experimental details related to the usage of idealized distribution prompts, examples of idealized distribution prompts used in~\Cref{sec:experimental_setup_idealized_distributions}, per task, as well as distribution description examples, and examples of few-shots. \newline
\noindent\textbf{\Cref{sec:appendix_idealized_distributions_results_summaries}} contains 3x4 summary figures corresponding to idealized distribution results described in~\Cref{sec:experimental_results_idealized_distributions}. \newline
\noindent\textbf{\Cref{sec:appendix_real_world_distribution_prompts}} contains examples of real-world distribution prompts used in~\Cref{sec:experimental_setup_real_world_distributions}. \newline
\noindent\textbf{\Cref{sec:appendix_additional_experimental_results}} contains additional model-wise experimental results that extend~\Cref{tab:model_comparison_RQ1} and~\Cref{tab:model_comparison_RQ2_RQ3}, normal approximation results that show whether or not invoking the true distribution name makes a difference, and results for Chain-of-Thought (CoT) and code tool-use. \newline
\noindent\textbf{\Cref{sec:appendix_broader_impacts}} contains our broader impacts statement.

%% file: p_appendix_sections/1-idealized_distribution_prompts.tex
\section{Idealized Distribution}
\label{sec:appendix_idealized_distribution_prompts}

\subsection{Experimental Details}
\label{sec:appendix_experimental_setup_idealized_distributions}

To systematically investigate performance on \emph{idealized distributions} using the three proposed tasks of \textit{estimating percentiles}, \textit{drawing samples}, and \textit{calculating probabilities}, our method involves sets of questions, or in the case of \textit{drawing samples}, a command, that systematically tests the model's knowledge of a given distribution. 
In addition to investigating our proposed tasks in a zero-shot setting, we consider two different ways of providing in-context examples (shots) in the prompt - \textit{within family shots} and \textit{within distribution shots}. 

\noindent\textbf{Zero-shot Performance.} We evaluate the zero-shot performance of three LMs (\emph{Gemini 1.0 Ultra}, \emph{GPT4-Turbo}, and \emph{GPT3.5-Turbo}) across our proposed tasks of \textit{estimating percentiles}, \textit{drawing samples}, and \textit{calculating probabilities} in order to evaluate an LM's understanding of probabilistic reasoning of the 12 idealized distributions described in~\Cref{sec:probabilistic_reasoning_datasets_idealized_distributions}. LM prompts are generated as formulated in~\Cref{sec:probabilistic_reasoning_tasks}.

\noindent\textbf{Performance by Shot Type.} We propose two shot types used across the aforementioned tasks - \textit{within distribution family distribution shots} and \textit{within distribution shots}. \textit{Within distribution family distribution shots} entail randomized shots from a different variant of the distribution being asked about in a question. The distribution parameters of the variant are randomly sampled from a specified range of reasonable parameter values. For example, if we are asking for a percentile of a value in a normal distribution with a mean of 100 and a standard deviation of 10, and we are providing three shots, the randomized shots may be generated from three variant normal distributions with means of 108, 118, and 112 and corresponding standard deviations of 13, 16, and 10. \textit{Within distribution shots} entail shots from the exact same distribution that is being asked about in a question. The shots are mapped per shot count to allow for a reasonable spread of shots throughout the distribution.

Additionally, for the task of \emph{estimating percentiles}, we compare both shot types to a baseline where the LM is asked to pick from one or more shots' answer based on the nearest corresponding target percentile value. This baseline represents a nearest neighbor approach using the set of in-context examples. This baseline makes appropriate use of the information given to the model, but importantly does not perform any interpolation between percentiles, which would be required for strong performance. If LM performance exceeded baseline performance, this would suggest that LMs perform some kind of interpolation, instead of simply reciting in-context examples. 

To avoid biasing our results, in the case of the \textit{estimating percentiles} and \textit{calculating probabilities} tasks, we explicitly avoid using shots that involve an answer (percentile or probability respectively) that could potentially be an answer to one of our proposed questions. Specifically, we use $n^{th}_{shots} = \{5, 15, 25, 35, 45, 55, 65, 75, 85, 95\}$ and $\mathcal{P}_{shots} = \{0.05, 0.15, 0.25, 0.35, 0.45, 0.55, 0.65, 0.75, 0.85, \newline0.95\}$. For example, if we are asking for a percentile of a value in a normal distribution with a mean of 100 and a standard deviation of 10, and we are providing three shots, the mapped shots will be generated from the same normal distribution and correspond to 35.0, 55.0, and 75.0. We sample \emph{Gemini 1.0 Ultra} responses 10 times per question with a random seed. Then, we elect to average and effectively use all answers as a part of our final evaluation in order to help capture the variability in language model responses and provide a broader understanding of their reasoning capabilities. This effectively is a form of self-consistency. We provide further details, including examples per shot type and shot count, in our~\Cref{sec:appendix}.

\noindent\textbf{Key LM Parameters for Reproducibility.} Language models typically have additional parameters, such as temperature and sampling strategies, which have default settings that can vary from model to model. For \emph{Gemini 1.0 Ultra}, \emph{GPT4-Turbo}, and \emph{GPT3.5-Turbo}, we utilize a default temperature of 0.7 for all of our experiments because we empirically discovered that a temperature of 0.7 to 0.9 with a random seed yielded similar, optimal performance on a hold-out dataset. A temperature of 0.7  also happens to be a default for many language models, such as various versions of the GPT and Gemini family models. Unless noted otherwise, results are obtained using a random seed.

Additional hyperparameters, such as frequency and brevity penalties, are not utilized in our experiments. We define a \emph{frequency penalty} as a penalty that is applied proportional to how many times a token has appeared in the response and prompt. We define a \emph{brevity penalty} as a penalty that targets responses that are very short, for example translations that contain only a few words. The nature of our proposed probabilistic reasoning questions, where the final answer from the instruction-tuned language model is always a single, numerical answer that is constrained by the prompt (for example, percentile answers of 25.3, 55.7, and 82.1) means that in both cases these penalties are not particularly relevant or effective at producing optimal answers. We utilized the default decoding approach for each model, with a default top-p (0.95), and where applicable, top-k (40).

\clearpage
\onecolumn

\subsection{Prompt Examples}

\vspace{0em}
\begin{tcolorbox}[colback=gray!5!white,colframe=gray!75!black,title=Percentile Example]
\footnotesize{
\#\# You are an expert on statistics. Your task is to estimate the percentile of a number within a specific distribution. Answer with just a numerical response from 0 to 100. Make sure your final answer is enclosed by xml tags <answer> and </answer>. \\

\#\# Here are some examples to help you understand the task: \\

\textbf{\{few\_shot\_examples\}} \\

\#\# Consider the following distribution: \\
\textbf{\{distribution\_description\}} \\

\#\# Here is your question: \\
Question: \\
What is the percentile of the value \textbf{\{target\_number\}} within the provided distribution? \\
Answer:
}
\end{tcolorbox}

\begin{tcolorbox}[colback=red!5!white,colframe=red!75!black,title=Sampling Example]
\footnotesize{
\#\# You are an expert on statistics. Your task is to sample a number from a given distribution. Do not write any code or use any additional tools to perform the sampling. Answer with just a numerical response. Make sure your final answer is enclosed by xml tags <answer> and </answer>. \\

\#\# Here are some examples to help you understand the task: \\

\textbf{\{few\_shot\_examples\}} \\

\#\# Consider the following distribution: \\
\textbf{\{distribution\_description\}} \\

\#\# Instruction: Sample a number from the given distribution and output only the numerical value.
}
\end{tcolorbox}

\begin{tcolorbox}[colback=blue!5!white,colframe=blue!75!black,title=Probability Example]
\footnotesize{
\#\# You are an expert on statistics. Your task is to estimate the probability of being in a range of values within a given distribution. Answer with just a numerical response from 0 to 1, representing the probability. Make sure your final answer is enclosed by xml tags <answer> and </answer>. \\

\#\# Here are some examples to help you understand the task: \\

\textbf{\{few\_shot\_examples\}} \\

\#\# Consider the following distribution: \\
\textbf{\{distribution\_description\}} \\

\#\# Here is your question: \\
Question: \\
Considering only values including and between the 1st percentile and the 99th percentile, what is the probability that a value from the provided distribution is between \textbf{\{lower\_target\_number\}} and \textbf{\{upper\_target\_number\}}? \\
Answer:
}
\end{tcolorbox}

\clearpage

\subsubsection{Distribution Description Examples}

\begin{tcolorbox}[colback=black!5!white,colframe=black!75!black,title=Normal]
\tiny{
Distribution Type: Normal Distribution\\
Mean: \textbf{\{mean\}}\\
Standard Deviation: \textbf{\{std\}}
}
\end{tcolorbox}

\begin{tcolorbox}[colback=black!5!white,colframe=black!75!black,title=Log-Normal]
\tiny{
Distribution Type: Log-Normal Distribution \\
Characteristics: This distribution models values that are the result of the multiplicative product of many independent random variables, such as income levels, stock prices, or city sizes
Log Mean (mu): \textbf{\{mean\}} \\
Log Sigma (sigma): \textbf{\{sigma\}} \\
These parameters mean that the natural logarithm of the values follows a normal distribution with the specified mean and standard deviation.
}
\end{tcolorbox}

\begin{tcolorbox}[colback=black!5!white,colframe=black!75!black,title=Exponential]
\tiny{
Distribution Type: Exponential Distribution \\
Characteristics: Models the time between events in a process where events occur continuously and independently at a constant average rate. \\
Rate: \textbf{\{rate\}} (The average number of events per unit time is \textbf{\{1/rate:.2f\}}.)
}
\end{tcolorbox}

\begin{tcolorbox}[colback=black!5!white,colframe=black!75!black,title=Power Law]
\tiny{
Distribution Type: Power Law Distribution \\
Characteristics: Known for its heavy tails suitable for describing phenomena with a high incidence of extreme values. \\
Alpha: \textbf{\{alpha\}} (Controls the tail heaviness—the smaller the alpha, the fatter the tail.) \\
Xmin: \textbf{\{xmin\}} (Minimum value for which the power law behavior holds.)
}
\end{tcolorbox}

\begin{tcolorbox}[colback=black!5!white,colframe=black!75!black,title=Uniform]
\tiny{
Distribution Type: Uniform Distribution \\
Characteristics: All values within the interval have equal probability of occurring. \\
Min: \textbf{\{a\}} (Minimum value of the distribution.) \\
Max: \textbf{\{b\}} (Maximum value of the distribution.)
}
\end{tcolorbox}

\begin{tcolorbox}[colback=black!5!white,colframe=black!75!black,title=Gamma]
\tiny{
Distribution Type: Gamma Distribution \\
Characteristics: Used to model waiting times and life data among other things. \\
Shape: \textbf{\{shape\}} (Controls the skewness of the distribution.) \\
Scale: \textbf{\{scale\}} (Controls the spread of the distribution.)
}
\end{tcolorbox}

\begin{tcolorbox}[colback=black!5!white,colframe=black!75!black,title=Skew-Normal]
\tiny{
Distribution Type: Skew-Normal Distribution \\
Characteristics: A generalization of the normal distribution to accommodate skewness. \\
Location: \textbf{\{location\}} (Shifts the distribution along the x-axis.) \\
Scale: \textbf{\{scale\}} (Controls the spread of the distribution.) \\
Skew: \textbf{\{skew\}} (Determines the direction and degree of skewness.)
}
\end{tcolorbox}

\begin{tcolorbox}[colback=black!5!white,colframe=black!75!black,title=Gumbel]
\tiny{
Distribution Type: Gumbel Distribution \\
Characteristics: Often used to model the distribution of extreme values. \\
Location: \textbf{\{loc\}} (Centers the distribution.) \\
Scale: \textbf{\{scale\}} (Controls the spread of the distribution.)
}
\end{tcolorbox}

\begin{tcolorbox}[colback=black!5!white,colframe=black!75!black,title=Poisson]
\tiny{
Distribution Type: Poisson Distribution \\
Characteristics: Suitable for modeling the number of events happening in a fixed interval of time or space. \\
Lambda: \textbf{\{lam\}} (Average rate of events per interval.)
}
\end{tcolorbox}

\begin{tcolorbox}[colback=black!5!white,colframe=black!75!black,title=Geometric]
\tiny{
Distribution Type: Geometric Distribution \\
Characteristics: Models the number of trials until the first success. \\
Probability of Success: \textbf{\{p\}}
}
\end{tcolorbox}

\begin{tcolorbox}[colback=black!5!white,colframe=black!75!black,title=Binomial]
\tiny{
Distribution Type: Binomial Distribution \\
Characteristics: Describes the number of successes in a fixed number of trials with a given probability of success. \\
Trials: \textbf{\{n\}} (Total number of trials.) \\
Probability of Success: \textbf{\{p\}} (Probability of success in each trial.)
}
\end{tcolorbox}

\begin{tcolorbox}[colback=black!5!white,colframe=black!75!black,title=Multinomial]
\tiny{
Distribution Type: Multinomial Distribution \\
Characteristics: Generalizes the binomial distribution for scenarios where each trial can result in more than two outcomes. \\
Trials: \textbf{\{n\}} (Total number of trials.) \\
Probabilities: \textbf{\{probs\}}
}
\end{tcolorbox}

\subsubsection{Examples of Few-shots}

\begin{tcolorbox}[colback=black!5!white,colframe=black!75!black,title=Within Family Shots]
Example 1:\\
Distribution:\\

  Distribution Type: Normal Distribution\\
  Mean: 80\\
  Standard Deviation: 10\\
  
Question:\\
What is the percentile of 74.722 within the provided distribution?\\
Answer:\\
<answer>30.0</answer>\\

Example 2:\\
Distribution:\\

  Distribution Type: Normal Distribution\\
  Mean: 108\\
  Standard Deviation: 13\\
  
Question:\\
What is the percentile of 107.903 within the provided distribution?\\
Answer:\\
<answer>50.0</answer>\\

Example 3:\\
Distribution:\\

  Distribution Type: Normal Distribution\\
  Mean: 82\\
  Standard Deviation: 8\\
  
Question:\\
What is the percentile of 88.708 within the provided distribution?\\
Answer:\\
<answer>80.0</answer>
\end{tcolorbox}

\begin{tcolorbox}[colback=black!5!white,colframe=black!75!black,title=Within Distribution Shots]
Example 1:\\
Distribution:\\

  Distribution Type: Normal Distribution\\
  Mean: 100\\
  Standard Deviation: 10\\
  
Question:\\
What is the percentile of 96.183 within the provided distribution?\\
Answer:\\
<answer>35.0</answer>\\

Example 2:\\
Distribution\\

  Distribution Type: Normal Distribution\\
  Mean: 100\\
  Standard Deviation: 10\\
  
Question:\\
What is the percentile of 101.298 within the provided distribution?\\
Answer:\\
<answer>55.0</answer>\\

Example 3:\\
Distribution:\\

  Distribution Type: Normal Distribution\\
  Mean: 100\\
  Standard Deviation: 10\\
  
Question:\\
What is the percentile of 106.802 within the provided distribution?\\
Answer:\\
<answer>75.0</answer>
\end{tcolorbox}

\twocolumn

%% file: p_appendix_sections/2-idealized_distribution_results_summaries.tex
\clearpage

\onecolumn
\section{Idealized Distributions Results Summaries}
\label{sec:appendix_idealized_distributions_results_summaries}

\begin{figure*}[ht!]
    \centering
    \includegraphics[width=1\textwidth]{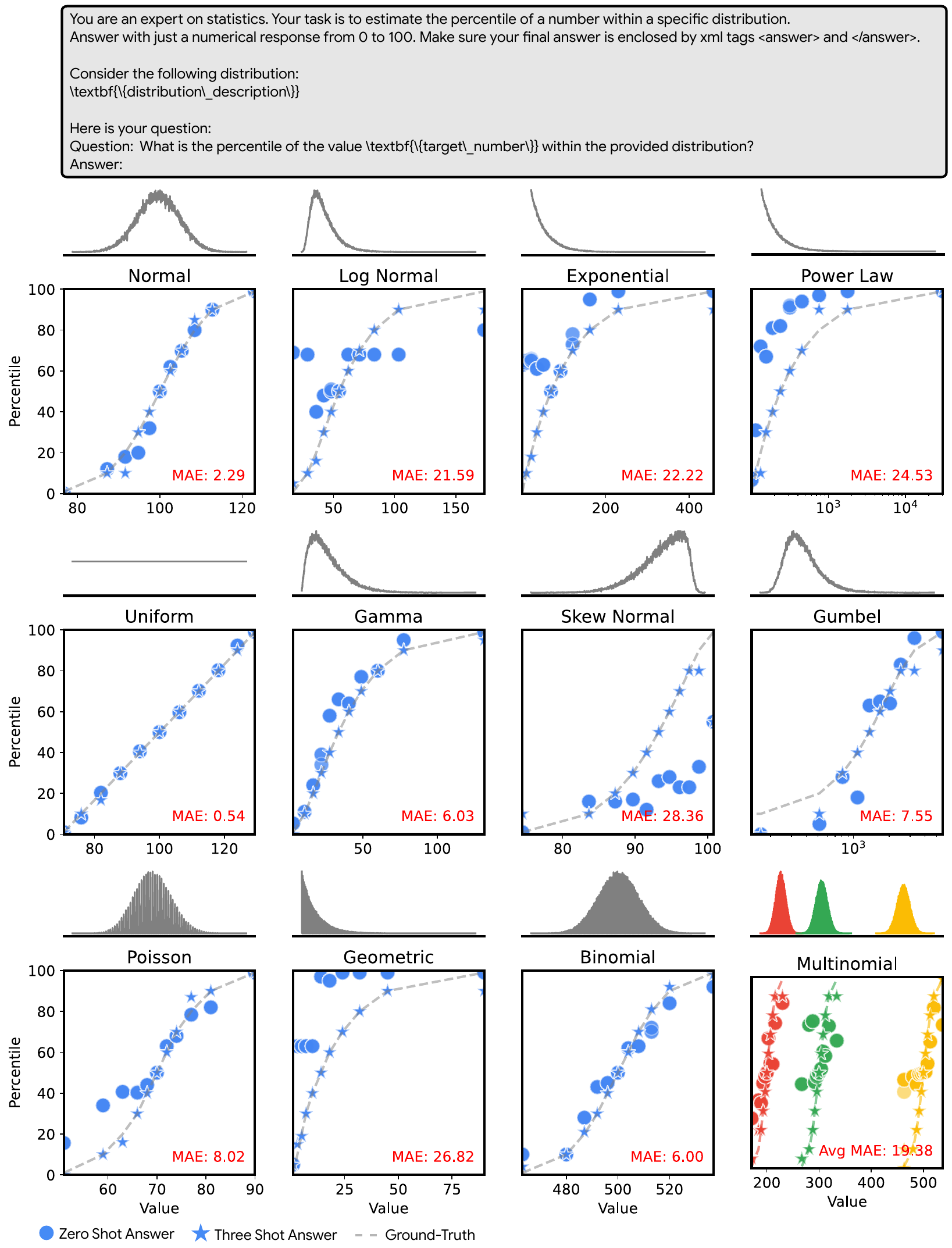}
    \caption{\textbf{Percentile Results.} Zero and three-shot (within distribution) results for \emph{returning percentile} estimations in each of the 12 families of distributions.}
    \label{fig:percentile_results_full}
\end{figure*}

\begin{figure*}[ht!]
    \centering
    \includegraphics[width=1\textwidth]{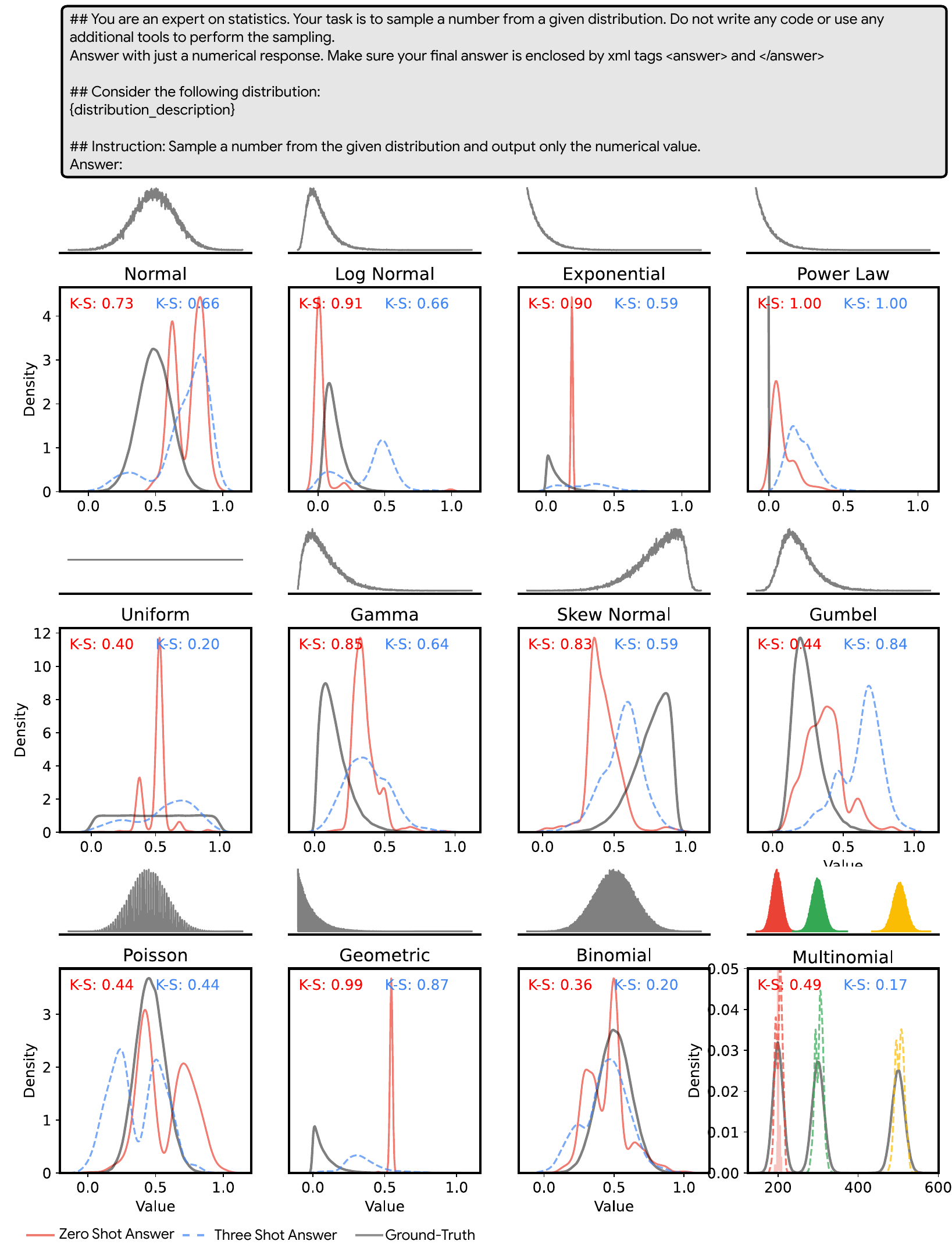}
    \caption{\textbf{Sampling Results.} Zero and three-shot (within distribution) results for \emph{drawing samples} (single repeated draws) in each of the 12 families of distributions.}
    \label{fig:sampling_results_full}
\end{figure*}

\begin{figure*}[ht!]
    \centering
    \includegraphics[width=1\textwidth]{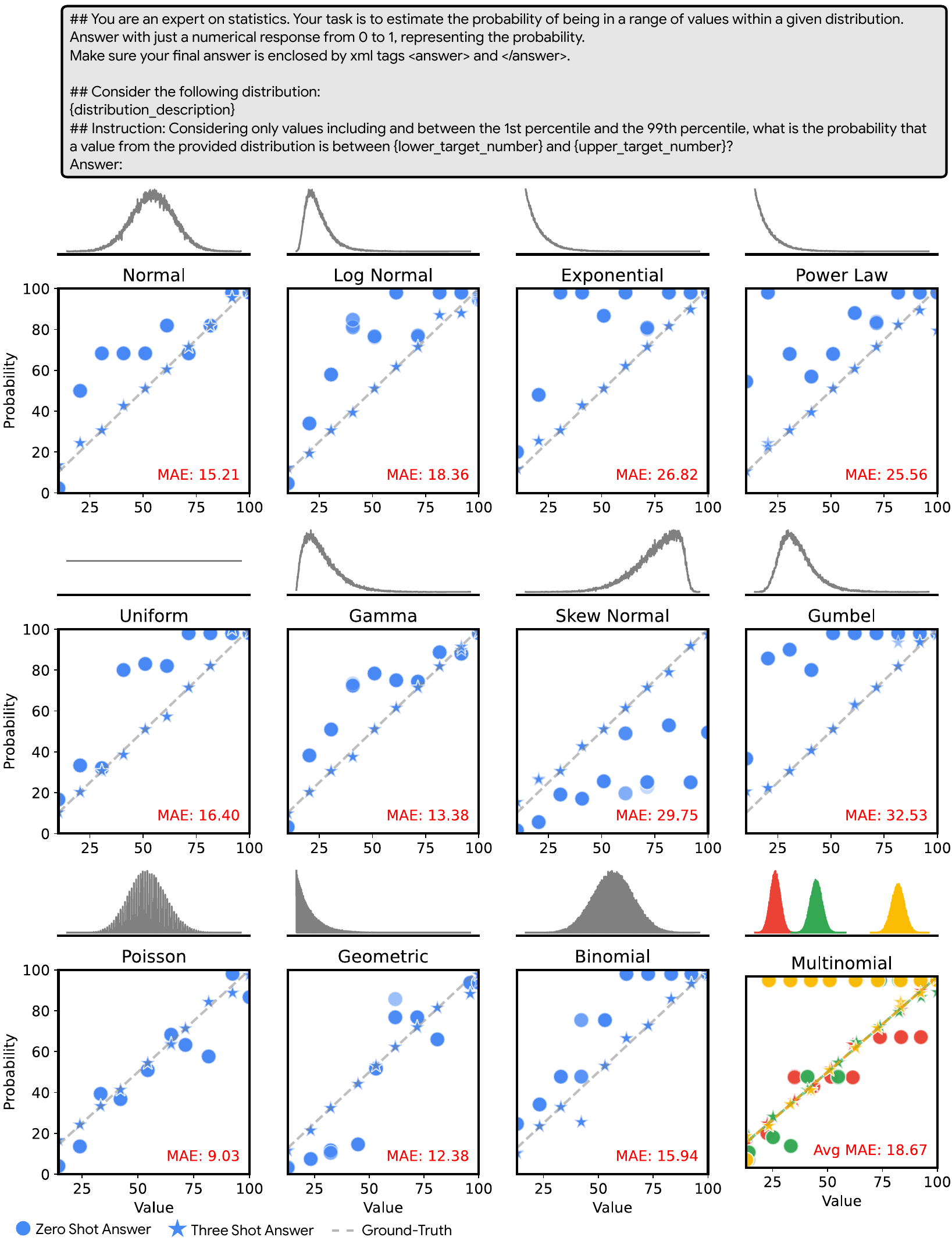}
    \caption{\textbf{Probability Results.} Zero and three-shot (within distribution) results for \emph{calculating probabilities} in each of the 12 families of distributions.}
    \label{fig:probabilities_results_full}
\end{figure*}
\twocolumn

%% file: p_appendix_sections/3-real_world_distribution_prompts.tex
\clearpage

\onecolumn
\section{Real-World Distribution Prompts}
\label{sec:appendix_real_world_distribution_prompts}

\begin{tcolorbox}[colback=gray!5!white,colframe=gray!75!black,title=Health Example]
\footnotesize{
\#\# You are an expert on population health and wearable fitness devices. Your task is to estimate the percentile of a given average step count value for a population that regularly uses Fitbit devices and is active on a daily basis. The data is filtered for individuals aged 18-65. The data is age-balanced and gender-balanced, and pertains to the U.S. population only. Do not use any additional tools such as code generation or search engines. Answer with just a numerical response from 0 to 100. Make sure your answer is enclosed by xml tags <answer> and </answer>. \\

\#\# Consider the following parameters that describe a normal distribution of this data: \\

  Mean: 8366.971 \\
  Standard Deviation: 3291.940 \\

\#\# Here is your question: \\
Question:\\
What is the percentile of the average step count value \textbf{\{target\_number\}} steps for users of Fitbit devices? Do not use any additional tools such as code generation or search engines. Answer with just a numerical response from 0 to 100. Make sure your answer is enclosed by xml tags <answer> and </answer>. \\
}
Answer:

\end{tcolorbox}

\begin{tcolorbox}[colback=red!5!white,colframe=red!75!black,title=Finance Example]
\footnotesize{
\#\# You are an expert on finance and statistics. Your task is to estimate the percentile of a given annual household income within the population using data from the year 2018 in the United States, sourced from the Census Bureau’s American Community Survey (ACS) Public Use Microdata Sample (PUMS). Do not use any additional tools such as code generation or search engines. Answer with just a numerical response from 0 to 100. Make sure your answer is enclosed by xml tags <answer> and </answer>. \\

\#\# Consider the following parameters that describe a Gumbel distribution of this data: \\

  Mean: 66028.713 \\
  Standard Deviation: 53616.018 \\

\#\# Here is your question: \\
Question: \\
What is the percentile of an annual household income value of \$\textbf{\{target\_number\}}? Do not use any additional tools such as code generation or search engines. Answer with just a numerical response from 0 to 100. Make sure your answer is enclosed by xml tags <answer> and </answer>. \\
Answer:
}
\end{tcolorbox}

\begin{tcolorbox}[colback=blue!5!white,colframe=blue!75!black,title=Climate Example]
\footnotesize{
\#\# You are an expert on climate science and statistics. Your task is to estimate the percentile of a given average temperature value using data from U.S. weather stations in the year 2018, sourced from the National Oceanic and Atmospheric Administration (NOAA) Global Historical Climatology Network Daily (GHCNd). Do not use any additional tools such as code generation or search engines. Answer with just a numerical response from 0 to 100. Make sure your answer is enclosed by xml tags <answer> and </answer>. \\

\#\# Consider the following parameters that describe a normal distribution of this data: \\

  Mean: 10.643 \\
  Standard Deviation: 12.628 \\

\#\# Here is your question:\\
Question: \\
What is the percentile of an average temperature of \textbf{\{targe\_number\}} degrees Celsius? Do not use any additional tools such as code generation or search engines. Answer with just a numerical response from 0 to 100. Make sure your answer is enclosed by xml tags <answer> and </answer>.\\
Answer:
}
\end{tcolorbox}

\twocolumn

%% file: p_appendix_sections/4-additional_experimental_results.tex
\section{Additional Experimental Results}
\label{sec:appendix_additional_experimental_results}

\subsection{Additional Model Results}
\label{sec:appendix_additional_model_results}

Additional model results, extending~\Cref{tab:model_comparison_RQ1} and~\Cref{tab:model_comparison_RQ2_RQ3}, can be found as a part of our GitHub repo here: \url{https://github.com/yahskapar/LLMs-and-Probabilistic-Reasoning}.

\subsection{CoT and Code Tool-use Results}
\label{sec:appendix_CoT_and_Code_Tool-use_results}

\begin{table*}[ht!]
    \normalsize	
    \captionsetup{width=\textwidth}
    \vspace{-8pt}
    \centering
    \normalsize	
    \setlength{\tabcolsep}{2pt}
    \adjustbox{max width=1\textwidth}{
    \begin{tabular}{rccccccccc}
    \toprule[1.5pt]
        & \multicolumn{3}{c}{\textbf{Health}} & \multicolumn{3}{c}{\textbf{Finance}} & \multicolumn{3}{c}{\textbf{Climate}} \\
        \cmidrule(lr){2-4} \cmidrule(lr){5-7} \cmidrule(lr){8-10}
        \textbf{Model} & \textbf{Norm. Approx.} & \textbf{CoT} & \textbf{RWC + Code} & \textbf{Norm. Approx.} & \textbf{CoT} & \textbf{RWC + Code} & \textbf{Norm. Approx.} & \textbf{CoT} & \textbf{RWC + Code} \\ 
        \midrule
        Gemini 1.0 Ultra & $\textbf{5.89} \pm \textbf{1.09}$ & $6.45 \pm 4.91$ & $6.62 \pm 1.03$ & $\textbf{7.62} \pm \textbf{1.06}$ & $9.45 \pm 1.26$ & $8.46 \pm 0.94$ & $\textbf{7.43} \pm \textbf{1.11}$ & $10.48 \pm 1.69$ & $8.56 \pm 0.92$ \\
        
        \bottomrule[1.5pt]
    \end{tabular}}
    \footnotesize
    \caption{\textbf{Zero-shot performance by domain and context category across different LMs.} RWC = Real-world Context. All results are reported as Mean Absolute Error (MAE) $\pm$ Standard Error ($\sigma_{M}$) with (\%) units.}
    \label{tab:model_comparison_CoT_and_Code}
    \vspace{-1.5em}
\end{table*}

\begin{figure*}[ht!]
    \centering
    \includegraphics[width=1\textwidth]{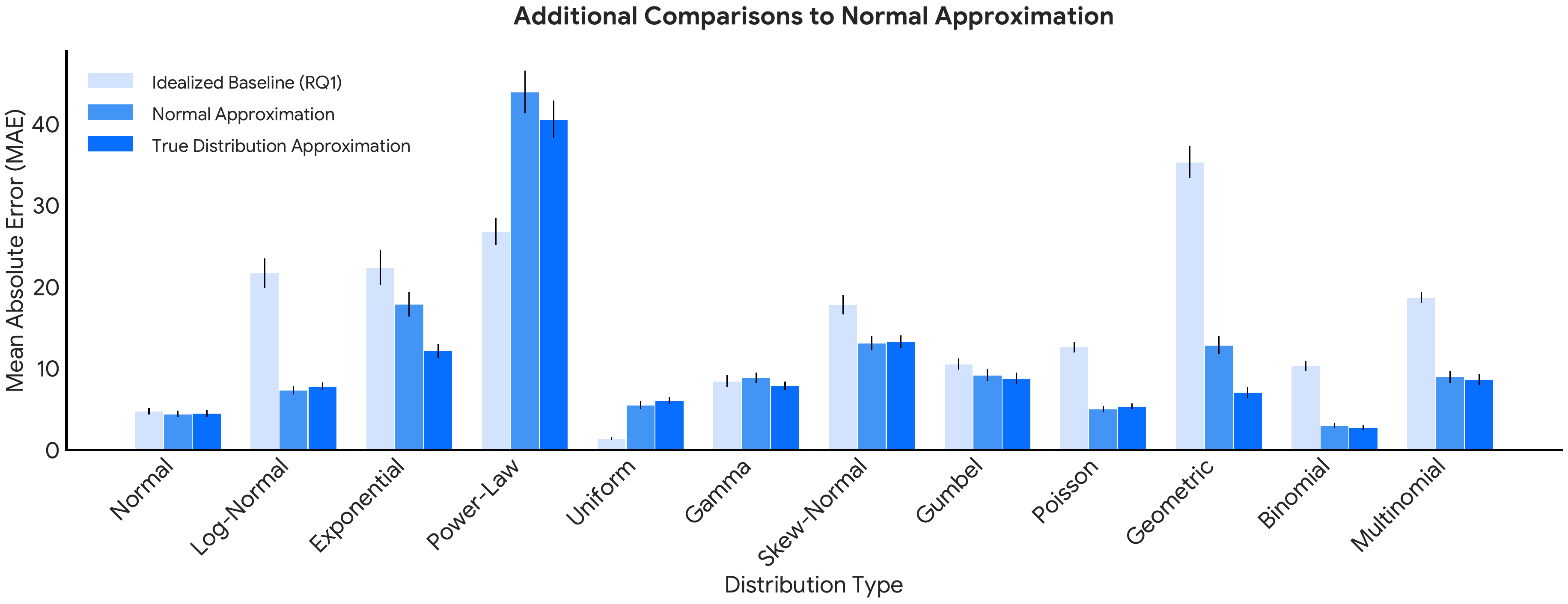}
    \vspace{-2em}
    \caption{\textbf{Additional Normal Approximation Results.} Additional idealized distribution results comparing the normal approximation approach to the baseline corresponding to idealized or real-world distributions and the true distribution approach.}
    \vspace{-1em}
    \label{fig:additional_normal_approximation_results}
\end{figure*}

We provide additional zero-shot Chain-of-Thought (CoT)~\cite{kojima2022large} and code tool-use results in~\Cref{tab:model_comparison_CoT_and_Code}. Though both CoT and code tool-use show benefits over assuming an idealized distribution or adding real-world context (\Cref{tab:model_comparison_RQ2_RQ3}), neither is convincingly better than the normal approximation approach. The fact that CoT results do not exceed normal approximation suggest that it is non-trivial to instruct a model to improve its modeling of non-normal distributions. Additional improvements to CoT-based approaches could be achieved with further investigation and development of techniques useful for numerical reasoning tasks. Furthermore, though we chose to use a cutoff data for corresponding datasets and did not employ a retrieval approach to retrieve and rank recent information that may be relevant to a proposed probabilistic reasoning question, we acknowledge that a retrieval tool can be useful to get more up-to-date information versus relying on parametric knowledge.

\subsection{Additional Normal Approximation Results}
\label{sec:appendix_additional_normal_approximation_results}

In~\Cref{fig:additional_normal_approximation_results}, we additionally present results that involve a variant of normal approximation where the true distribution is assumed with a mean and standard deviation as a part of the prompt. In the case of idealized distributions, we utilize the same distribution name with provided mean and standard deviation. In the case of real-world distributions, we approximate the distribution based on the K-S statistic after a matching process across all 12 idealized distributions described in~\Cref{sec:probabilistic_reasoning_datasets_idealized_distributions}.

%% file: p_appendix_sections/5-broader_impacts.tex
\section{Broader Impacts}
\label{sec:appendix_broader_impacts}

Though we pose more constrained, systematic questions as a part of our investigation of language models and their ability to perform probabilistic reasoning, real-world questions such as "is taking 8000 steps a day normal for an average adult in the U.S. population using wearable devices?" and others that we pose as a part of~\autoref{sec:probabilistic_reasoning_datasets_real_world_distributions} are completely reasonable questions for an average user of LMs to ask. There is a significant practical impact in improving the probabilistic reasoning capabilities of language models on real-world distributions, especially when answers to the aforementioned reasonable questions can affect a user's perception of real-world distributions and ultimately their perspective on potentially critical matters such as those in the domains of health, finance, and climate. 

%% file: main.bbl
\begin{thebibliography}{38}
\providecommand{\natexlab}[1]{#1}

\bibitem[{Achiam et~al.(2023)Achiam, Adler, Agarwal, Ahmad, Akkaya, Aleman,
  Almeida, Altenschmidt, Altman, Anadkat et~al.}]{achiam2023gpt}
Josh Achiam, Steven Adler, Sandhini Agarwal, Lama Ahmad, Ilge Akkaya,
  Florencia~Leoni Aleman, Diogo Almeida, Janko Altenschmidt, Sam Altman,
  Shyamal Anadkat, et~al. 2023.
\newblock Gpt-4 technical report.
\newblock \emph{arXiv preprint arXiv:2303.08774}.

\bibitem[{Bachmann and Nagarajan(2024)}]{bachmann2024pitfalls}
Gregor Bachmann and Vaishnavh Nagarajan. 2024.
\newblock The pitfalls of next-token prediction.
\newblock \emph{arXiv preprint arXiv:2403.06963}.

\bibitem[{Chakravarti et~al.(1967)Chakravarti, Laha, and
  Roy}]{chakravarti1967handbook}
Indra~Mohan Chakravarti, Radha~Govira Laha, and Jogabrata Roy. 1967.
\newblock Handbook of methods of applied statistics.
\newblock \emph{Wiley Series in Probability and Mathematical Statistics (USA)
  eng}.

\bibitem[{Chater et~al.(2006)Chater, Tenenbaum, and
  Yuille}]{chater2006probabilistic}
Nick Chater, Joshua~B Tenenbaum, and Alan Yuille. 2006.
\newblock Probabilistic models of cognition: Conceptual foundations.
\newblock \emph{Trends in cognitive sciences}, 10(7):287--291.

\bibitem[{Chen et~al.(2023)Chen, Wong, Chen, and Tian}]{chen2023extending}
Shouyuan Chen, Sherman Wong, Liangjian Chen, and Yuandong Tian. 2023.
\newblock Extending context window of large language models via positional
  interpolation.
\newblock \emph{arXiv preprint arXiv:2306.15595}.

\bibitem[{Chen et~al.(2022)Chen, Ma, Wang, and Cohen}]{chen2022program}
Wenhu Chen, Xueguang Ma, Xinyi Wang, and William~W Cohen. 2022.
\newblock Program of thoughts prompting: Disentangling computation from
  reasoning for numerical reasoning tasks.
\newblock \emph{arXiv preprint arXiv:2211.12588}.

\bibitem[{Cobbe et~al.(2021)Cobbe, Kosaraju, Bavarian, Chen, Jun, Kaiser,
  Plappert, Tworek, Hilton, Nakano et~al.}]{cobbe2021training}
Karl Cobbe, Vineet Kosaraju, Mohammad Bavarian, Mark Chen, Heewoo Jun, Lukasz
  Kaiser, Matthias Plappert, Jerry Tworek, Jacob Hilton, Reiichiro Nakano,
  et~al. 2021.
\newblock Training verifiers to solve math word problems.
\newblock \emph{arXiv preprint arXiv:2110.14168}.

\bibitem[{Cosmides and Tooby(1996)}]{cosmides1996humans}
Leda Cosmides and John Tooby. 1996.
\newblock Are humans good intuitive statisticians after all? rethinking some
  conclusions from the literature on judgment under uncertainty.
\newblock \emph{cognition}, 58(1):1--73.

\bibitem[{Frank(2009)}]{frank2009common}
Steven~A Frank. 2009.
\newblock The common patterns of nature.
\newblock \emph{Journal of evolutionary biology}, 22(8):1563--1585.

\bibitem[{Gao et~al.(2023)Gao, Madaan, Zhou, Alon, Liu, Yang, Callan, and
  Neubig}]{gao2023pal}
Luyu Gao, Aman Madaan, Shuyan Zhou, Uri Alon, Pengfei Liu, Yiming Yang, Jamie
  Callan, and Graham Neubig. 2023.
\newblock Pal: Program-aided language models.
\newblock In \emph{International Conference on Machine Learning}, pages
  10764--10799. PMLR.

\bibitem[{Geva et~al.(2020)Geva, Gupta, and Berant}]{geva2020injecting}
Mor Geva, Ankit Gupta, and Jonathan Berant. 2020.
\newblock Injecting numerical reasoning skills into language models.
\newblock In \emph{Proceedings of the 58th Annual Meeting of the Association
  for Computational Linguistics}, pages 946--958.

\bibitem[{Goldstein and Rothschild(2014)}]{goldstein2014lay}
Daniel~G Goldstein and David Rothschild. 2014.
\newblock Lay understanding of probability distributions.
\newblock \emph{Judgment and Decision making}, 9(1):1--14.

\bibitem[{He-Yueya et~al.(2023)He-Yueya, Poesia, Wang, and
  Goodman}]{he2023solving}
Joy He-Yueya, Gabriel Poesia, Rose~E Wang, and Noah~D Goodman. 2023.
\newblock Solving math word problems by combining language models with symbolic
  solvers.
\newblock \emph{arXiv preprint arXiv:2304.09102}.

\bibitem[{Imani et~al.(2023)Imani, Du, and Shrivastava}]{imani2023mathprompter}
Shima Imani, Liang Du, and Harsh Shrivastava. 2023.
\newblock Mathprompter: Mathematical reasoning using large language models.
\newblock \emph{arXiv preprint arXiv:2303.05398}.

\bibitem[{Knill and Pouget(2004)}]{knill2004bayesian}
David~C Knill and Alexandre Pouget. 2004.
\newblock The bayesian brain: the role of uncertainty in neural coding and
  computation.
\newblock \emph{TRENDS in Neurosciences}, 27(12):712--719.

\bibitem[{Kojima et~al.(2022)Kojima, Gu, Reid, Matsuo, and
  Iwasawa}]{kojima2022large}
Takeshi Kojima, Shixiang~Shane Gu, Machel Reid, Yutaka Matsuo, and Yusuke
  Iwasawa. 2022.
\newblock Large language models are zero-shot reasoners.
\newblock \emph{Advances in neural information processing systems},
  35:22199--22213.

\bibitem[{Lewkowycz et~al.(2022)Lewkowycz, Andreassen, Dohan, Dyer,
  Michalewski, Ramasesh, Slone, Anil, Schlag, Gutman-Solo
  et~al.}]{lewkowycz2022solving}
Aitor Lewkowycz, Anders Andreassen, David Dohan, Ethan Dyer, Henryk
  Michalewski, Vinay Ramasesh, Ambrose Slone, Cem Anil, Imanol Schlag, Theo
  Gutman-Solo, et~al. 2022.
\newblock Solving quantitative reasoning problems with language models.
\newblock \emph{Advances in Neural Information Processing Systems},
  35:3843--3857.

\bibitem[{Lindskog et~al.(2021)Lindskog, Nystr{\"o}m, and
  Gredeb{\"a}ck}]{lindskog2021can}
Marcus Lindskog, P{\"a}r Nystr{\"o}m, and Gustaf Gredeb{\"a}ck. 2021.
\newblock Can the brain build probability distributions?
\newblock \emph{Frontiers in Psychology}, 12:596231.

\bibitem[{Liu et~al.(2022)Liu, Liu, Lu, Welleck, West, Le~Bras, Choi, and
  Hajishirzi}]{liu2022generated}
Jiacheng Liu, Alisa Liu, Ximing Lu, Sean Welleck, Peter West, Ronan Le~Bras,
  Yejin Choi, and Hannaneh Hajishirzi. 2022.
\newblock Generated knowledge prompting for commonsense reasoning.
\newblock In \emph{Proceedings of the 60th Annual Meeting of the Association
  for Computational Linguistics (Volume 1: Long Papers)}, pages 3154--3169.

\bibitem[{Liu et~al.(2024)Liu, Wu, Wu, Lu, Chang, and Feng}]{liu2024llms}
Xiao Liu, Zirui Wu, Xueqing Wu, Pan Lu, Kai-Wei Chang, and Yansong Feng. 2024.
\newblock Are llms capable of data-based statistical and causal reasoning?
  benchmarking advanced quantitative reasoning with data.
\newblock \emph{arXiv preprint arXiv:2402.17644}.

\bibitem[{Lu et~al.(2022)Lu, Qiu, Yu, Welleck, and Chang}]{lu2022survey}
Pan Lu, Liang Qiu, Wenhao Yu, Sean Welleck, and Kai-Wei Chang. 2022.
\newblock A survey of deep learning for mathematical reasoning.
\newblock \emph{arXiv preprint arXiv:2212.10535}.

\bibitem[{McDuff et~al.(2023)McDuff, Schaekermann, Tu, Palepu, Wang, Garrison,
  Singhal, Sharma, Azizi, Kulkarni et~al.}]{mcduff2023towards}
Daniel McDuff, Mike Schaekermann, Tao Tu, Anil Palepu, Amy Wang, Jake Garrison,
  Karan Singhal, Yash Sharma, Shekoofeh Azizi, Kavita Kulkarni, et~al. 2023.
\newblock Towards accurate differential diagnosis with large language models.
\newblock \emph{arXiv preprint arXiv:2312.00164}.

\bibitem[{Menne et~al.(2012)Menne, Durre, Vose, Gleason, and
  Houston}]{menne2012overview}
Matthew~J Menne, Imke Durre, Russell~S Vose, Byron~E Gleason, and Tamara~G
  Houston. 2012.
\newblock An overview of the global historical climatology network-daily
  database.
\newblock \emph{Journal of atmospheric and oceanic technology}, 29(7):897--910.

\bibitem[{Merrill et~al.(2024)Merrill, Paruchuri, Rezaei, Kovacs, Perez, Liu,
  Schenck, Hammerquist, Sunshine, Tailor, Ayush, Su, He, McLean, Malhotra,
  Patel, Zhan, Althoff, McDuff, and Liu}]{merrill2024transforming}
Mike~A. Merrill, Akshay Paruchuri, Naghmeh Rezaei, Geza Kovacs, Javier Perez,
  Yun Liu, Erik Schenck, Nova Hammerquist, Jake Sunshine, Shyam Tailor, Kumar
  Ayush, Hao-Wei Su, Qian He, Cory~Y. McLean, Mark Malhotra, Shwetak Patel,
  Jiening Zhan, Tim Althoff, Daniel McDuff, and Xin Liu. 2024.
\newblock \href {https://arxiv.org/abs/2406.06464} {Transforming wearable data
  into health insights using large language model agents}.
\newblock \emph{Preprint}, arXiv:2406.06464.

\bibitem[{Oaksford and Chater(2001)}]{oaksford2001probabilistic}
Mike Oaksford and Nick Chater. 2001.
\newblock The probabilistic approach to human reasoning.
\newblock \emph{Trends in cognitive sciences}, 5(8):349--357.

\bibitem[{Ozturkler et~al.(2023)Ozturkler, Malkin, Wang, and
  Jojic}]{ozturkler2023thinksum}
Batu Ozturkler, Nikolay Malkin, Zhen Wang, and Nebojsa Jojic. 2023.
\newblock Thinksum: Probabilistic reasoning over sets using large language
  models.
\newblock In \emph{Proceedings of the 61st Annual Meeting of the Association
  for Computational Linguistics (Volume 1: Long Papers)}, pages 1216--1239.

\bibitem[{Ruggles et~al.(2020)Ruggles, Flood, Goeken, Grover, Meyer, Pacas, and
  Sobek}]{ruggles2020ipums}
Steven Ruggles, Sarah Flood, Ronald Goeken, Josiah Grover, Erin Meyer, Jose
  Pacas, and Matthew Sobek. 2020.
\newblock Ipums usa: version 10.0 [dataset].
\newblock \emph{Minneapolis, Mn: Ipums}, 10:D010.

\bibitem[{Saxton et~al.(2019)Saxton, Grefenstette, Hill, and
  Kohli}]{saxton2019analysing}
David Saxton, Edward Grefenstette, Felix Hill, and Pushmeet Kohli. 2019.
\newblock Analysing mathematical reasoning abilities of neural models.
\newblock \emph{arXiv preprint arXiv:1904.01557}.

\bibitem[{Shaham et~al.(2022)Shaham, Segal, Ivgi, Efrat, Yoran, Haviv, Gupta,
  Xiong, Geva, Berant et~al.}]{shaham2022scrolls}
Uri Shaham, Elad Segal, Maor Ivgi, Avia Efrat, Ori Yoran, Adi Haviv, Ankit
  Gupta, Wenhan Xiong, Mor Geva, Jonathan Berant, et~al. 2022.
\newblock Scrolls: Standardized comparison over long language sequences.
\newblock \emph{arXiv preprint arXiv:2201.03533}.

\bibitem[{Tang et~al.(2023)Tang, Sun, Idnay, Nestor, Soroush, Elias, Xu, Ding,
  Durrett, Rousseau et~al.}]{tang2023evaluating}
Liyan Tang, Zhaoyi Sun, Betina Idnay, Jordan~G Nestor, Ali Soroush, Pierre~A
  Elias, Ziyang Xu, Ying Ding, Greg Durrett, Justin~F Rousseau, et~al. 2023.
\newblock Evaluating large language models on medical evidence summarization.
\newblock \emph{npj Digital Medicine}, 6(1):158.

\bibitem[{Team et~al.(2023)Team, Anil, Borgeaud, Wu, Alayrac, Yu, Soricut,
  Schalkwyk, Dai, Hauth et~al.}]{team2023gemini}
Gemini Team, Rohan Anil, Sebastian Borgeaud, Yonghui Wu, Jean-Baptiste Alayrac,
  Jiahui Yu, Radu Soricut, Johan Schalkwyk, Andrew~M Dai, Anja Hauth, et~al.
  2023.
\newblock Gemini: a family of highly capable multimodal models.
\newblock \emph{arXiv preprint arXiv:2312.11805}.

\bibitem[{Touvron et~al.(2023)Touvron, Martin, Stone, Albert, Almahairi,
  Babaei, Bashlykov, Batra, Bhargava, Bhosale et~al.}]{touvron2023llama}
Hugo Touvron, Louis Martin, Kevin Stone, Peter Albert, Amjad Almahairi, Yasmine
  Babaei, Nikolay Bashlykov, Soumya Batra, Prajjwal Bhargava, Shruti Bhosale,
  et~al. 2023.
\newblock Llama 2: Open foundation and fine-tuned chat models.
\newblock \emph{arXiv preprint arXiv:2307.09288}.

\bibitem[{Tversky and Kahneman(1974)}]{tversky1974judgment}
Amos Tversky and Daniel Kahneman. 1974.
\newblock Judgment under uncertainty: Heuristics and biases: Biases in
  judgments reveal some heuristics of thinking under uncertainty.
\newblock \emph{science}, 185(4157):1124--1131.

\bibitem[{Webb et~al.(2023)Webb, Holyoak, and Lu}]{webb2023emergent}
Taylor Webb, Keith~J Holyoak, and Hongjing Lu. 2023.
\newblock Emergent analogical reasoning in large language models.
\newblock \emph{Nature Human Behaviour}, 7(9):1526--1541.

\bibitem[{Workshop et~al.(2022)Workshop, Scao, Fan, Akiki, Pavlick, Ili{\'c},
  Hesslow, Castagn{\'e}, Luccioni, Yvon et~al.}]{workshop2022bloom}
BigScience Workshop, Teven~Le Scao, Angela Fan, Christopher Akiki, Ellie
  Pavlick, Suzana Ili{\'c}, Daniel Hesslow, Roman Castagn{\'e}, Alexandra~Sasha
  Luccioni, Fran{\c{c}}ois Yvon, et~al. 2022.
\newblock Bloom: A 176b-parameter open-access multilingual language model.
\newblock \emph{arXiv preprint arXiv:2211.05100}.

\bibitem[{Yang et~al.(2022)Yang, Qin, Chen, Lin, and
  Liang}]{yang2022logicsolver}
Zhicheng Yang, Jinghui Qin, Jiaqi Chen, Liang Lin, and Xiaodan Liang. 2022.
\newblock Logicsolver: Towards interpretable math word problem solving with
  logical prompt-enhanced learning.
\newblock In \emph{Findings of the Association for Computational Linguistics:
  EMNLP 2022}, pages 1--13.

\bibitem[{Zhang et~al.(2024{\natexlab{a}})Zhang, Da, Lee, Robinson, Wu, Song,
  Zhao, Raja, Slack, Lyu et~al.}]{zhang2024careful}
Hugh Zhang, Jeff Da, Dean Lee, Vaughn Robinson, Catherine Wu, Will Song,
  Tiffany Zhao, Pranav Raja, Dylan Slack, Qin Lyu, et~al. 2024{\natexlab{a}}.
\newblock A careful examination of large language model performance on grade
  school arithmetic.
\newblock \emph{arXiv preprint arXiv:2405.00332}.

\bibitem[{Zhang et~al.(2024{\natexlab{b}})Zhang, Ladhak, Durmus, Liang,
  McKeown, and Hashimoto}]{zhang2024benchmarking}
Tianyi Zhang, Faisal Ladhak, Esin Durmus, Percy Liang, Kathleen McKeown, and
  Tatsunori~B Hashimoto. 2024{\natexlab{b}}.
\newblock Benchmarking large language models for news summarization.
\newblock \emph{Transactions of the Association for Computational Linguistics},
  12:39--57.

\end{thebibliography}
